\documentclass[runningheads]{llncs}
\usepackage[T1]{fontenc}
\usepackage{graphicx}
\usepackage{booktabs}
\usepackage{amsmath}
\usepackage{amssymb}
\usepackage{multirow}
\usepackage{color, xcolor}
\usepackage{comment}
\usepackage{tcolorbox}
\usepackage{mathtools}

\usepackage{graphicx}    
\usepackage[hidelinks]{hyperref}    
\usepackage{subcaption} 

\usepackage[ruled,vlined]{algorithm2e}

\hypersetup{
    colorlinks=true,
    linkcolor=blue,
    urlcolor=blue,
}

\usepackage[top=2.5cm, bottom=2.5cm, left=3.2cm, right=3.2cm]{geometry}
\usepackage[utf8]{inputenc}

\usepackage{graphicx} 
\usepackage{fancyhdr} 

\usepackage{booktabs} 

\makeatletter
\newcommand{\cdotfill}{%
    \leavevmode \cleaders \hb@xt@ .5em{\hss $\cdot$ \hss}\hfill \kern 0pt %
}
\makeatother

\begin{document}
%

%
%

%
%

%
\authorrunning{F. Author et al.}
%
\institute{Tencent, Beijing, China \\
\email{\{wendongbi\}@tencent.com}}

\quad 

\quad

\begin{center}
    {\LARGE\bfseries WeMusic-Agent: Efficient Conversational Music Recommendation via Knowledge Internalization and Agentic Boundary Learning} \\[1.2cm]
    { Wendong Bi\textsuperscript{1}, Yirong Mao\textsuperscript{1}, Xianglong Liu\textsuperscript{2}, Kai Tian\textsuperscript{3}, Jian Zhang\textsuperscript{1}, Hanjie Wang\textsuperscript{1}, Wenhui Que\textsuperscript{1}\footnote{Corresponding Author}} \\[0.3cm]
    $\prescript{1}{}{\text{WeChat, Tencent Inc, Beijing, China}}$\qquad$\prescript{2}{}{\text{Peking University}}$ \qquad $\prescript{3}{}{\text{Tsinghua University}}$ \\[0.3cm]
    { \{wendongbi, erongmao, torchzhang, hankinwang, victorque\}@tencent.com} \\[0.1cm]
    {liuxianglong@pku.edu.cn} \\[0.1cm]
     {tk23@mails.tsinghua.edu.cn }\\[0.3cm]
    
    
    \includegraphics[height=1.0em]{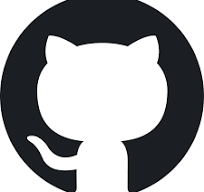}~
    \href{https://github.com/wemusicmodel/WeMusicAgent}{\texttt{https://github.com/wemusicmodel/WeMusicAgent}}
\end{center}
%
%


\begin{abstract}

Personalized music recommendation in conversational scenarios usually requires a deep understanding of user preferences and nuanced musical context, yet existing methods often struggle with balancing specialized domain knowledge and flexible tool integration. This paper proposes WeMusic-Agent, a training framework for efficient LLM-based conversational music recommendation. By integrating the knowledge internalization and agentic boundary learning, the framework aims to teach the model to intelligently decide when to leverage internalized knowledge and when to call specialized tools (e.g., music retrieval APIs, music recommendation systems). Under this framework, we present WeMusic-Agent-M1, an agentic model that internalizes extensive musical knowledge via continued pretraining on 50B music-related corpus while acquiring the ability to invoke external tools when necessary. Additionally, considering the lack of open-source benchmarks for conversational music recommendation, we also construct a benchmark for personalized music recommendations derived from real-world data in WeChat Listen. This benchmark enables comprehensive evaluation across multiple dimensions, including relevance, personalization, and diversity of the recommendations. Experiments on real-world data demonstrate that WeMusic-Agent achieves significant improvements over existing models. 

\end{abstract}



\newpage 

\section*{Contents} 
\label{sec:contents} 


\begin{enumerate}
    \item \hyperlink{sec:chapter_1_intro}{\textcolor{black}{\textbf{Introduction}}} \cdotfill \textbf{3}\\
    \begin{itemize}
        \item[] \hyperlink{sec:chapter_1_1}{\textcolor{black}{1.1 Contributions}} \cdotfill 4\\
        \item[] \hyperlink{sec:chapter_1_2}{\textcolor{black}{1.2 Overview of WeMusic-Agent}} \cdotfill 4\\
    \end{itemize}
    \vspace{6pt}
    
    \item \hyperlink{sec:chapter_2_wemusic_base}{\textcolor{black}{\textbf{WeMusic-Base: Internalizing Music Knowledge into LLMs}}} \cdotfill \textbf{5}\\
    \begin{itemize}
        \item[] \hyperlink{sec:chapter_2_2}{\textcolor{black}{2.1 Continual Pretraining on Large-scale Music Corpus}} \cdotfill 5\\
        \item[] \hyperlink{sec:chapter_2_3}{\textcolor{black}{2.2 Supervised Finetuning}} \cdotfill 8\\
        \item[] \hyperlink{sec:chapter_2_4}{\textcolor{black}{2.3 Multi-Objective Rewards for Reinforcement Learning}} \cdotfill 9\\
        \item[] \hyperlink{sec:chapter_2_5}{\textcolor{black}{2.4 Self-Distillation: From Single-song Recommendation to List-wise
Recommendation}} \cdotfill 11\\
    \end{itemize}
    \vspace{6pt}
    
    \item \hyperlink{sec:chapter_3_wemusic_agent}{\textcolor{black}{\textbf{WeMusic-Agent: Breaking Through the Boundary of Internalization}}} \cdotfill \textbf{11}\\
    \begin{itemize}
        \item[] \hyperlink{sec:chapter_3_1}{\textcolor{black}{3.1 Internalized Model vs. Agent Model: Is There A Golden Rule in Music CRS?}} \cdotfill 11\\
        \item[] \hyperlink{sec:chapter_3_2}{\textcolor{black}{3.2 Overview of WeMusic-Agent}} \cdotfill 11\\
        \item[] \hyperlink{sec:chapter_3_3}{\textcolor{black}{3.3 WeMusic-Agent-Zero: Music Agentic Model with Tool Calling for Cold Start}} \cdotfill 13\\
        \item[] \hyperlink{sec:chapter_3_4}{\textcolor{black}{3.4 Agentic Trajectory Sampling: Discovering the Boundary of Internalization}} \cdotfill 13\\
        \item[] \hyperlink{sec:chapter_3_5}{\textcolor{black}{3.5 WeMusic-Agent-M1: Efficient Agentic Boundary Learning}} \cdotfill 13\\
    \end{itemize}
    \vspace{6pt}
    
    \item \hyperlink{sec:chapter_4_benchmark}{\textcolor{black}{\textbf{WeMusic-Bench: Benchmark and Metrics for Music Recommendation}}} \cdotfill \textbf{14}\\
    \vspace{6pt}
    
    \item \hyperlink{sec:chapter_5_exp}{\textcolor{black}{\textbf{Experiments and Discussion}}} \cdotfill \textbf{15}\\
    \begin{itemize}
        \item[] \hyperlink{sec:chapter_5_1}{\textcolor{black}{5.1 Main Results}} \cdotfill 15\\
        \item[] \hyperlink{sec:chapter_5_2}{\textcolor{black}{5.2 Discussions on The Effects of MusicCPT}} \cdotfill 17\\
        
        \item[] \hyperlink{sec:chapter_5_4}{\textcolor{black}{5.3 Effectiveness of Diversity Rewards in WeMusic-Base}} \cdotfill 18\\
        \item[] \hyperlink{sec:chapter_5_3}{\textcolor{black}{5.4 Discussions on Rewards and Tool Efficiency}} \cdotfill 18\\
    \end{itemize}
    \vspace{6pt}
    
    \item \hyperlink{sec:chapter_6_conclusion}{\textcolor{black}{\textbf{Conclusion, Limitations and Future Work}}} \cdotfill \textbf{19}\\
    \vspace{6pt}
    
    \item \hyperlink{sec:chapter_7_disclaimer}{\textcolor{black}{\textbf{Disclaimer and Security Analysis}}} \cdotfill \textbf{19}\\
    \vspace{6pt}

    \item[] \hyperlink{sec:reference}{\textcolor{black}{\textbf{References}}} \cdotfill \textbf{19}\\
    \vspace{6pt}

    \item[] \hyperlink{sec:appendix}{\textcolor{black}{\textbf{Appendix}}} \cdotfill \textbf{21}\\
    \vspace{6pt}
    
\end{enumerate}
\newpage

\section{Introduction}
\hypertarget{sec:chapter_1_intro}{}

Conversational recommendation systems (CRS)\cite{sun2018conversational} have emerged as a transformative paradigm in AI-driven music recommendation, offering dynamic interaction capabilities that transcend the limitations of traditional music recommendation systems \cite{Jannach_2021,palumbo2025text2trackspromptbasedmusicrecommendation,afchar2022explainability}. 

In recent years, the rapid development of large language models has driven the evolution of Conversational Music Recommendation \cite{doh2025talkplay}. However, existing general-purpose large language models usually do not have enough knowledge in the area of music, and effective music recommendation in the conversational scenario remains a challenging problem. Existing methods of domain knowledge injection for language models mainly include knowledge internalization methods \cite{TalkPlay,melchiorre2025just,palumbo2025text2tracks} and agent methods \cite{doh2025talkplay,11199092,11174114,AgentCF}. Knowledge internalization methods train the model with large-scale data to internalize music knowledge into the parameters of the model during the training process. Although injecting a large amount of knowledge into the model at once may help, it still faces issues such as outdated information, hallucinations, and limited knowledge coverage. Recently, agent methods greatly solve the problems of capability boundaries and hallucinations by introducing external tools and interacting with the environment in real time. However, it is time-consuming in conversational music recommendation tasks and lacks the ability to understand personalized user preferences. 

In this paper, we propose WeMusic-Agent, a novel training framework to integrate the leverage of internalized knowledge and agentic tools naturally. This framework includes two pipelines, i.e., a multi-stage knowledge internalization pipeline (WeMusic-Base) and an agentic boundary learning pipeline (WeMusic-Agent). WeMusic-Base continually pretrained on the Qwen2.5-32B \cite{qwen2,qwen3} Model and post-trained for single-song recommendation in multi-turn dialogue scenarios. We first perform continual pretraining via a dual reference model framework on the 50B token-level music corpus to inject music knowledge into the model (MusicCPT) without catastrophic forgetting in general knowledge. And then we conduct supervised fine‑tuning in multi-turn dialogue scenarios with generated data. Finally, we further design a multi-objective reinforcement learning algorithm to further improve the performance of music recommendation and align the human preferences. Specifically, we design a hybrid reward function for multi-objective optimization with GRPO \cite{shao2024deepseekmathpushinglimitsmathematical} that considers relevance, personalization, diversity, and factuality. And we observe that the reward normalization strategy is helpful for balanced optimization and stabilizing the RL process. And the resulting model WeMusic-Base 32B gains significant improvements in music recommendation tasks  without external knowledge compared to other general-purpose LLMs of hundreds of billions of parameters. Besides, to enable song-list recommendation, we also design a self-distillation algorithm by repeated sampling from WeMusic-Base to get a list of songs as the answer for each query. 

With more internalized music knowledge, the internalized models, e.g., WeMusic-Base, have better adaptation for user preference, a more timely and fluent dialogue experience in conversational music recommendation scenario. However, they may suffer from weak timeliness and hallucinations for out-of-knowledge questions, thus have explict capability boundary. So we propose WeMusic-Agent, an agentic model trained from WeMusic-Base-Dist, which calls external music searching tools when necessary.  We first conduct agentic RL on WeMusic-Base-Dist to get a naive agentic model named WeMusic-Agent-Zero that responds with tool calling. Then we conduct Agentic Trajectory Sampling on WeMusic-Base-Dist and WeMusic-Agent-Zero simultaneously. Then we define the capability of the knowledge internalization by positive samples (in-knowledge) and negative samples (out-of-knowledge) from WeMusic-Base-Dist. And we collect the positive samples of WeMusic-Base-Dist and replace the negative samples with agnetic samples from WeMusic-Agent-Zero as the data for agentic boundary tuning in a curriculum learning manner. Next, we design a controllable reinforcement learning algorithm to learn the agentic boundary. Finally, we get WeMusic-Agent-M1, which responds directly to in-knowledge queries while calling the search tools first before responding to out-of-knowledge queries.

To address the lack of open-source benchmarks for conversational music recommendation, we collect a large amount of real-world user queries in Chinese from WeChat Listen and then propose a benchmark for conversational music recommendation. Besides, we also design multi-dimensional evaluation metrics to evaluate the performance of CRS models in tasks of music recommendation and dialogues. Comprehensive experiments on real-world benchmarks indicate the excellent performance of WeMusic-Agent on conversational music recommendation tasks compared with the SOTA open-source and closed-source LLMs.

\subsection{Contributions}
\hypertarget{sec:chapter_1_1}{}
\subsubsection{A Multi-stage Knowledge Internalization Framework for LLM-based Music Recommendation} \quad\\
We propose WeMusic, a training framework that internalizes comprehensive music knowledge, as well as our base model for WeMusic-Agent.
\begin{itemize}
    \item Music Continual Pretraining at Scale.
    \item Post-Training on Multi-turn Conversational Music Recommendation Tasks.
    \item A Multi-objectives Reinforcement Learning Algorithm for Personalized Music Recommendation.
\end{itemize}

\subsubsection{Agentic Boundary Learning: Balancing the efficiency and effectiveness for utilization of internal knowledge and external tools.}

\begin{itemize}
    \item Self-Distillation from single-song recommendation to playlist recommendation.
    \item WeMusic-Agent: we propose the Agentic Boundary Learning framework, including the Curriculum Learning Guided SFT and Controllable Reinforcement Learning.
\end{itemize}

\subsubsection{An industry-level benchmark on conversational music recommendation.}
\begin{itemize}
    \item Data Collection: we collected comprehensive real-world music search and conversational data in Chinese from WeChat Listen, including both precise music searches (e.g., song, singer) and fuzzy music searches (e.g., style, scenario, genre, emotion).
    \item Evaluation Metrics: we introduce a multi-dimensional evaluation framework to evaluate the performance of music recommendations via relevance, personalization and diversity.
\end{itemize}





\subsection{Overview of WeMusic-Agent}
\begin{figure*}[h]
\includegraphics[width=\textwidth]{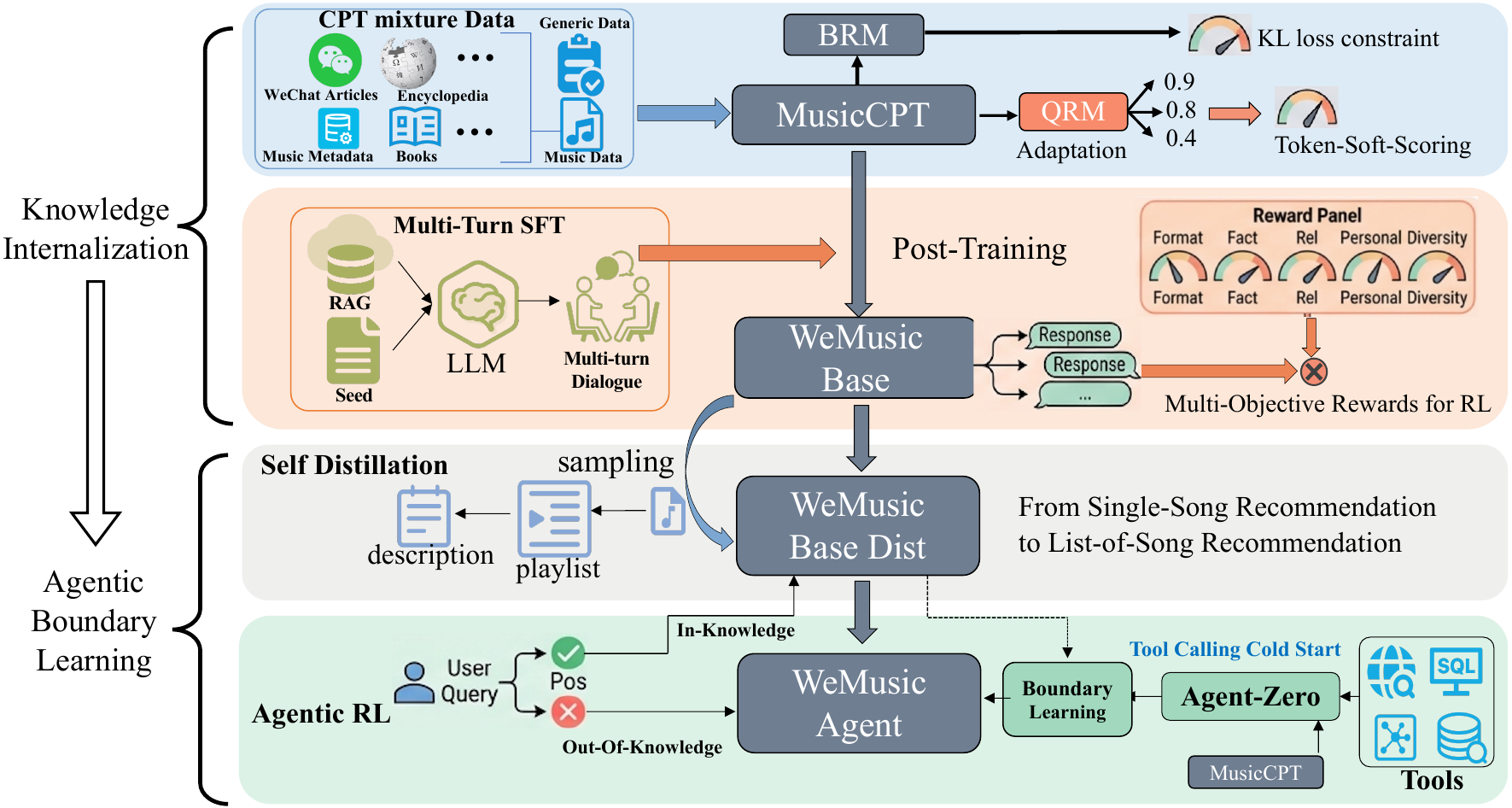} 
\caption{An overview of the framework of WeMusic-Agent.}
\label{fig:music_agent_overview}
\end{figure*}

\hypertarget{sec:chapter_1_2}{}
WeMusic is designed to support personalized music interactions, including song search, recommendation, and conversational engagement. As shown in the Figure~\ref{fig:music_agent_overview}, we present an overview of WeMusic, including the training pipeline for WeMusic-Base and WeMusic-Agent. Specifically, we first conduct music knowledge internalization based on open-sourced LLMs (Qwen2.5), including continual pretraining and post-training. And then we further conduct Agentic Boundary Learning to get our WeMusic-Agent model with necessary tool calling.

\label{sec:format_overview}

To train the model, we first develop a structured instruction template that guides the base model to follow specific response guidelines.
Given an input containing dialogue history, user query, and contextual information (e.g., currently playing song, time), the output is formatted according to the following template:
\begin{equation}
    \nonumber
    \begin{aligned}
        &\texttt{<intention>} \text{intention of user} \texttt{</intention>}\\
        &\texttt{<music>} \text{[\{"song\_name": xxx, "singer\_name": xxx\}...]} \texttt{</music>}\\
        &\texttt{<text>} \text{description of recommendations} \texttt{</text>}
    \end{aligned}
\end{equation}

The output template comprises three structured components:
1) The \texttt{<intention>} field captures the system’s predicted intent based on the user’s prompt, such as “music search” or “chat”.
2) The \texttt{<music>} component provides song recommendations, formatted as a list containing song names and corresponding singer names. It supports both single-song recommendations and playlist-style recommendations.
3) The \texttt{<text>} field contains the natural-language description of the recommended songs, tailored to the context of the user’s query.

\section{WeMusic-Base: Internalizing Music Knowledge into LLMs}
\hypertarget{sec:chapter_2_wemusic_base}{}
We first introduce WeMusic-Base, the model with large-scale music knowledge internalization, as the base model for WeMusic-Agent.

\subsection{Continual Pretraining on Large-scale Music Corpus}
\hypertarget{sec:chapter_2_2}{}
\label{sec:cpt}
To facilitate large-scale Continual Pre-Training (\textbf{CPT}), we first construct a comprehensive music corpus characterized by its scale, quality, and diversity. This corpus encompasses multiple forms of textual data, including song-related articles, encyclopedia, emotional and contextual annotations, user comments, playlist, books, and song metadata and etc, laying a robust groundwork for systematically internalizing music knowledge. On the training front, we introduce a Dual Reference Model framework. The first reference model assigns fine-grained, token-level quality scores to samples in the music-domain corpus, which are subsequently utilized to reweight the continual pre-training loss. Drawing inspiration from the concept of lifelong learning without forgetting in continual training, as evidenced in related literature \cite{li2017learning,distilqwen,baichuan_finance}, the second reference model constrains the model’s output distribution on general-domain data, thereby mitigating catastrophic forgetting as the model specializes toward the music domain. In what follows, we first detail the construction of our music-domain dataset, and followed by a comprehensive presentation of our Dual Reference Model framework for continual pre-training.

To construct the music-domain corpus, we adopt a document selection strategy focusing on strong relevance to songs, artists, and authentic listening behaviors, sourcing from large-scale generic texts and content within the WeChat ecosystem. We first train a lightweight domain classifier on high-quality music-related articles alongside song and artist descriptions to facilitate this selection. Subsequently, we implement a multi-stage cleaning pipeline comprising language normalization, quality scoring, template paragraph removal, near-duplicate elimination, and masking of potentially privacy-sensitive spans. This pipeline maintains the overall scale of the corpus while enhancing its purity and reducing noise that could hinder the convergence of continual pre-training. Following iterative rounds of filtering, cleaning, and resampling, we obtain a refined music-domain corpus comprising approximately 50 billion tokens, designated for continual pre-training. The details of the data construction of CPT process can be found in the appendix~\ref{sec:cpt_data_appendix}.
\begin{figure*}[h]
\centering
\includegraphics[width=1.01\textwidth]{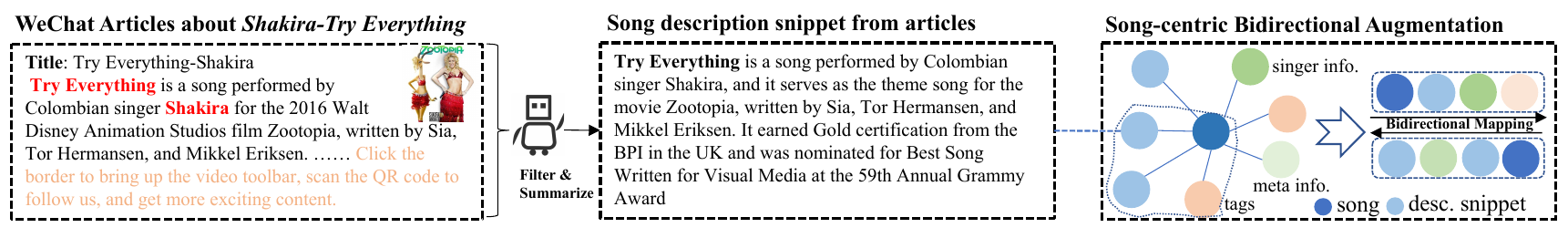} 
\caption{Song description snippet mining and song-centric bidirectional augmenation pipeline. First, articles and comments of the same song are filtered and summarized into high quality snippets by DeepSeek-V3 \cite{deepseek_v3}. Then song-related information including the above description snippets, singer-related information, tags, and etc are linked into song-centric graph where subgraph with multiple nodes are sampled to synthesize song-to-description and description-to-song training data.}
\label{fig:bi_aug}
\end{figure*}

\textbf{Song-Centric Bidirectional Data Augmentation.} In human-written articles or encyclopedia pages to describe songs, the song name usually appears first, followed by an introduction to its related description. If trained on such data with next token prediction (\textbf{NTP}), models generally become proficient at generating descriptive content from a given song name, due to the fact that current prevalent decode-only LLM utilize left-to-right attention mechanisms. However, for query-based song recommendation task, one of the most important tasks of our model, requires the model to have the ability to predict song name from relevant descriptive information. Thus, the NTP attention orders in articles/encyclopedia pages and query-based song recommendation task are different: the first is song-to-description, and the latter is description-to-song. Furthermore, there is a reversal curse that models trained on $A \mapsto B$ mapping often struggle with $B \mapsto A$ \cite{reversal_curse}. Based on such insights, we augmentation bidirectional mapping data to enhance the song search capability as shown in Fig.~\ref{fig:bi_aug}. 

\textbf{Dual Reference Model framework.} We follow the standard autoregressive language modeling objective: given a training sample \(x = (x_1, \dots, x_T)\), the model is optimized by minimizing the average cross entropy over all tokens,
\[
\mathcal{L}_{\text{CPT}} = - \mathbb{E}_{x}\Big[\sum_{t=1}^{T} \log p_\theta(x_t \mid x_{<t})\Big].
\]
Under this formulation, all tokens contribute equally to the objective. However, in real world music domain corpora, there is a large variance in both quality and information density across segments: fine-grained descriptions of song emotion and style, or users' detailed reflections on a melody, are typically far more valuable for making the model "understand music" than templated copyright notices, platform boilerplate, or noisy advertisement words. Simply running equal weight CPT on the raw corpus tends to waste capacity on low-value tokens and makes the model more susceptible to noise, which in turn hurts the efficiency of injecting music knowledge.
\begin{figure*}[h]
\centering
\includegraphics[width=\textwidth]{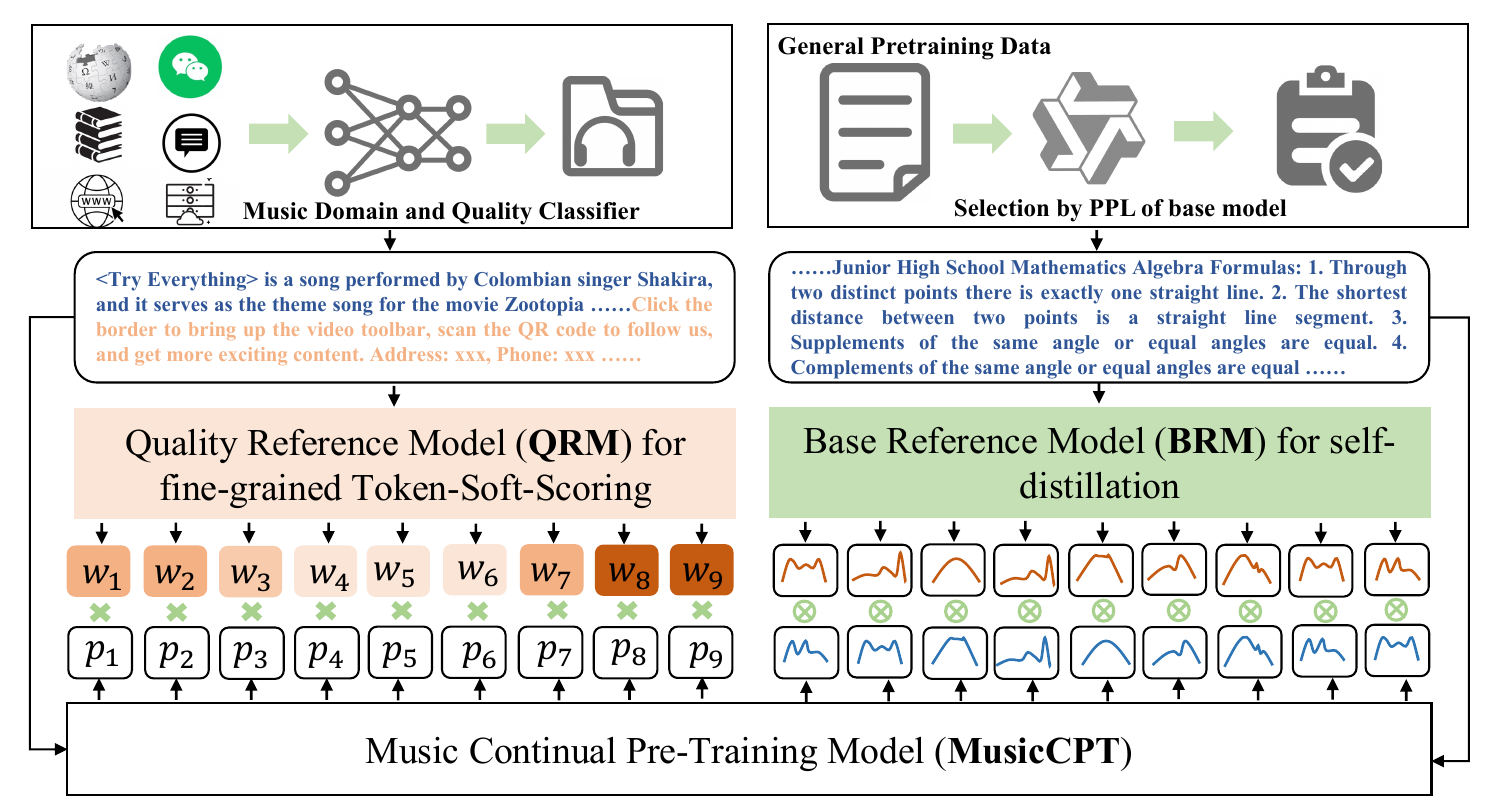} 
\caption{Music domain Continual Pre-Training(MuCPT) method overview}
\label{fig:cpt_overview}
\end{figure*}

To address this issue, we introduce a music reference model, namely the Quality Reference Model (\textbf{QRM}), which assigns fine-grained token level scores for quality and domain relevance, and uses them to re-weight the CPT loss on music domain data. Concretely, for each sequence labeled as music domain, we first use the QRM to predict a scalar score for every token, indicating its importance along the dimensions of music relevance and information quality. We rewrite the CPT objective on the music corpus as
\[
\mathcal{L}_{\text{Music}} = - \mathbb{E}_{x \in \mathcal{D}_{\text{music}}}\Big[\sum_{t=1}^{T} w_t \log p_\theta(x_t \mid x_{<t})\Big].
\]
In this way, tokens belonging to high quality song reviews, professional critiques, and style/emotion descriptions are amplified during training, while templated phrases, and noisy content are relatively suppressed. At the same time, we do not perform hard truncation or removal on the original text as in \cite{lin2024rho,prox_example}; the model still models the full context, striking a balance between preserving semantic coherence and increasing the intellectual density of the training signal. 

In implementation, \(w_t\) is not a manually tuned hyper parameter or annotated by Super LLM \cite{fineweb,qurating} but is directly induced from the token-level loss of the reference model which is trained over high quality seed samples with next token prediction. We first compute the token-level negative log-likelihood of the music reference model
\[
C_{\text{QRM}}(x_t) = -\log p_{\text{QRM}}(x_t \mid x_{<t}),
\]
and then define
\[
w_t = \frac{1}{C_{\text{QRM}}(x_t)}.
\]
In other words, tokens that the reference model finds easy to predict (lower loss, higher probability belonging to distribution of high quality data) receive larger normalized weights, while high-loss tokens are softly down-weighted.

In addition to the music reference model, we introduce a general reference model, namely the Base Reference Model (\textbf{BRM}), to explicitly regularize the model's general domain knowledge. The BRM is not an extra, larger teacher model; instead, it is simply the open source base model at the checkpoint before continual pretraining starts. In this way, we can inject the constraint of not drifting too far from the original general capability into the training process via self-distillation, without adding any extra model parameters or engineering complexity.

Specifically, for samples in general domain, we compute both the student distribution \(p_\theta(x_t \mid x_{<t})\) and the BRM distribution \(p_{\text{BRM}}(x_t \mid x_{<t})\) at each position, and minimize the KL divergence between their logits based output distributions, forming a self-distillation loss:
\[
\mathcal{L}_{\text{General}} = \mathbb{E}_{x \in \mathcal{D}_{\text{general}}}\Big[\sum_{t=1}^{T} \mathrm{KL}\big(p_{\text{BRM}}(\cdot \mid x_{<t}) \,\|\, p_\theta(\cdot \mid x_{<t})\big)\Big].
\]
In practice, \(\mathcal{L}_{\text{General}}\) can be linearly combined with the standard language modeling loss so that, while the model absorbs information from a small amount of new domain data, it is also pulled back towards the original base model, preventing noticeable degradation in general semantics, world knowledge, and instruction following ability.

\textbf{Two Stage Continual Pre-training Strategy.}
In \textbf{Stage-1}, we continue to pre-train models with the mixture of generic and music domain data where the music domain data is about 50B tokens in total.
For \textbf{Stage-2}, after the initial stage, the model has been injected with wide and substantial music-related knowledge; however, its memory for song names remains insufficiently robust, leading it to frequently hallucinate fake song names. Thus, we increase the proportion of song-related tokens such as song comments, descriptive snippets, QA-like song search data and etc in this stage.

\subsection{Supervised Fine-tuning}
\hypertarget{sec:chapter_2_3}{}
To equip the CPT model with foundational multi‑turn dialogue capability and exposure to varied conversational patterns, we first construct a supervised fine‑tuning (SFT) dataset and subsequently fine‑tune the model on it. More concretely, we design modality‑aware data synthesis pipelines to generate high‑quality training samples tailored to different data types. Following this, we adopt a curriculum‑style SFT strategy that incrementally trains the model, first on conversational skills and later on specialized recommendation tasks.

\subsubsection{Data Generation.}
To cover diverse conversational formats and topics, we create synthetic data as a fundamental resource for SFT. These data broadly fall into two categories: pattern-oriented and capability-oriented. Pattern-oriented data capture 10 common human–assistant interaction templates, exposing the model to varied conversational styles, user intents, and linguistic habits. Emphasizing structural diversity, they include dialogues such as greetings, clarifications, preference expressions, and casual inquiries, enabling the model to generalize across heterogeneous user behaviors and adapt to varied dialogue patterns in practice. Capability-oriented data, in contrast, aim to strengthen the model's core conversational abilities. These samples are categorized into 8 classes, targeting higher-level skills including multi‑turn memory tracking, coreference resolution, coherent context propagation, safety‑aware response generation, and principled handling of refusal or uncertainty. By distilling such competencies, this category enhances the reliability, contextual consistency, and overall user experience of the conversational recommendation system.
Our synthetically generated dialogues are structured as multi-turn exchanges spanning 3–10 conversational turns. Each dialogue instance is organized into three fundamental segments: \textbf{system}, \textbf{human}, and \textbf{gpt}. The system part contains the system prompt and an optional user profile, which includes demographic attributes (age, gender, education level, city) alongside music engagement history, specifically sets of songs the user has liked and completed. The human part encodes the user's active playback context (currently playing song), underlying intent, and the detailed textual query. The gpt part reflects the predicted user intent, the model's response, and the recommended song(s) when the predicted intent corresponds to song search. Supported user intents cover four main categories: (1) non-music chat (e.g., casual conversation), (2) music-related chat, (3) music recommendation, and (4) playback control (e.g., start, skip/next, thumbs-up, single-track repeat). 
Leveraging the WeChat Listen platform’s Top-140K popular tracks, we assemble a seed corpus for synthetic dialogue generation. This corpus integrates multiple facets of music-related metadata, including track titles, artist names, song narratives, user commentary, and biographical profiles of artists.
Overall, we employ three distinct synthesis strategies to construct the SFT dataset, as detailed below.
\begin{enumerate}
  \item \textbf{N-RAG-based method}. To ensure coherence in multi-turn music-related dialogues, we synthesize the majority of the data through an N-RAG pipeline. First, we randomly sample $n$ songs from the CPT corpus and prompt an LLM to extract multiple thematic clusters from their associated metadata and textual descriptions. For each identified theme, an RAG module fetches the $m$ most relevant songs based on semantic relevance. The aggregated metadata and descriptive texts of these songs are then concatenated and input to the DeepSeek-V3 model to generate synthetic multi-turn dialogues. This RAG-enhanced approach maintains high consistency within the seed corpus, resulting in synthetic conversations that are both natural and linguistically fluent. Specially, by providing topic words and complexity levels, we drive the model to produce a diverse set of song search queries, and apply generated verification conditions to remove low-quality instances. Those queries are sampled and utilized in multiple stages including CPT, SFT and RAG tool.
  \begin{figure*}[h]
    \centering
    \includegraphics[width=0.90\textwidth]{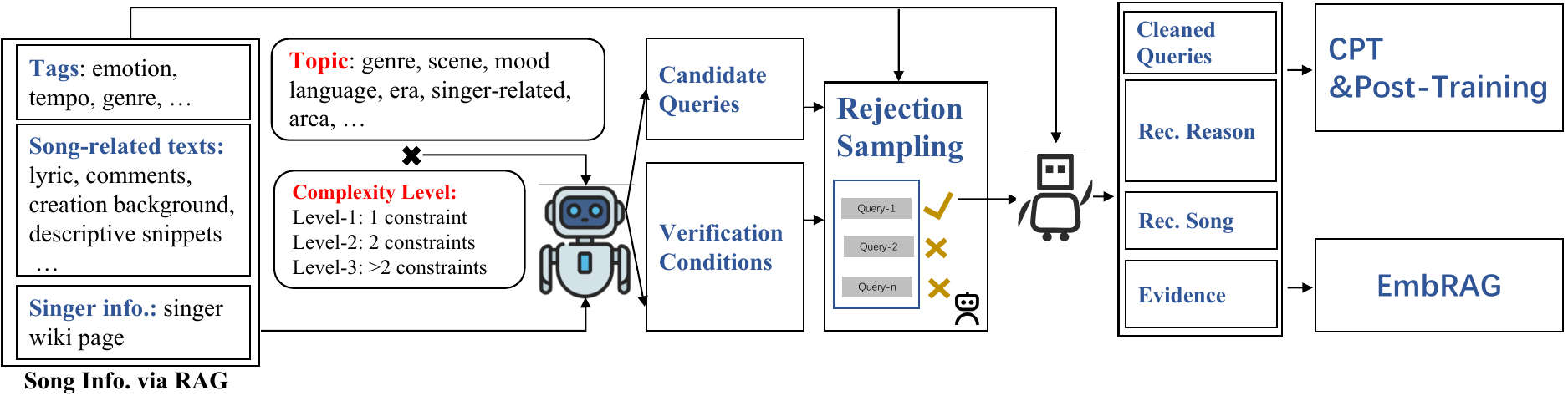} 
    \caption{Pipeline to synthesize diversity song search queries.}
    \label{fig:song_search_qa_aug}
    \end{figure*}
  \item \textbf{Seed-library-based method}. For synthetic subsets that demand high diversity such as capability-boundary data, we adopt a seed-library-based method. Concretely, we curate a seed library of heterogeneous data templates, uniformly sample templates from this library, and input both the sampled templates and the raw corpus into DeepSeek-V3 for diverse data generation. Relying solely on the template set in the prompt tends to induce homogenized generations.
  \item \textbf{Meta-prompt-based method}. We annotated a small amount of pattern-oriented examples manually, and devised a meta-prompt–based method to generate more similar data. Concretely, we supply an LLM with the few-shot examples and the pattern type, instructing it to write a meta prompt that describes how to synthesize the data; this generated meta prompt is then concatenated with a generic prompt to form the final instruction. This method dynamically adapts the synthesis process to sparse data and specific requirements, thereby significantly minimizing the time and human effort involved.
\end{enumerate}

\subsubsection{Supervised Fine-Tuning}
Through supervised fine-tuning on this dataset, the model develops multi‑turn dialogue capability and learns to model personalized associations between users and content.
The input $x$ comprises the system prompt $s$, user profile $u$, dialogue history $t$, and current query $q$. The output $y$ consists of the textual response and the recommended music. The cross-entropy loss function in the SFT stage is:

\begin{equation}
\begin{aligned}
 \mathcal{L}_{\text{SFT}}
&= - \mathbb{E}_{(x,y)\sim\mathcal{D_{SFT}}}
\left[
    \sum_{t=1}^{T} \log p_{\theta}\bigl(y_t \mid x, y_{<t}\bigr)
\right] \\
x &= \left[s,u,t,q\right]
\end{aligned}
\end{equation}

To ensure a high-quality model distribution during SFT, we adopt a curriculum-like training method that mitigates the risk that large volumes of synthetic data dilute the contribution of high-quality human-curated data.
\begin{enumerate}
  \item \textbf{Early stage} Before the learning rate reaches its maximum, we oversample high-quality human-curated data to anchor the model in a strong initial distribution.
  \item \textbf{Middle stage} We fine-tune on a mixture of filtered synthetic and human data. This enables the model to learn diverse dialog patterns, but may introduce distributional drift due to the limited fidelity of synthetic data.

  \item \textbf{Late stage} As the learning rate enters the annealing phase, we again fine-tune with oversampled human data to pull the learned distribution back toward the target.
\end{enumerate}
This method maximizes the impact of human-curated data while drawing on the broad coverage provided by synthetic samples, thereby enhancing the model’s alignment with the human-expected conversational distribution and maintaining high-quality interactions throughout the SFT process.

\subsection{Multi-Objective Rewards for Reinforcement Learning}
\hypertarget{sec:chapter_2_4}{}
\label{sec:rl_internal}

To address various requirements, we designed corresponding rewards and constructed a hybrid Multi-Reward system.  Given the conversational context history \( H \), the model generates a set of responses $\overline{A}=\{A_i\}_L=\{(T_i,M_i)\}_L$, where \( L \) represents the number of responses generated per context (we set $L=8$ in our experiments), $T_i$ and $M_i$ represent the output text (\texttt{<text>} part) and recommended song (\texttt{<music>} part) of $A_i$ (see Section~\ref{sec:format_overview}). Then the Reward is computed as follows:

\subsubsection{Format Reward}
To ensure the model's responses adhere to the standard format illustrated in Section~\ref{sec:format_overview}. The melody is refreshing, and the lyrics also have a very summery vibe~</text>\textbackslash n <music>Song: Common Jasmine Orange\textbackslash nArtist: Jay Chou</music>'). We employ regular expressions to match the model's output and assign corresponding format rewards based on the degree of format compliance \(R_{\text{Format}}(A_i)\in\{0,0.5,1\}\).

\subsubsection{Factuality Reward}\label{fact}
To mitigate the issue of entity hallucination in music recommendation models, we have constructed and maintained a database containing approximately 3 million authentic music entity records. For each recommended entity \(M_i\) generated by the model, the system performs a matching query against this database: if \(M_i\) exists, it is classified as a valid entity; otherwise, it is deemed a hallucinated entity. Based on this verification mechanism, we define a binary factuality reward function \(R_\text{Factuality} \in\{0,1\}\), which is utilized during training and evaluation to constrain the authenticity of entities generated by the model.

\subsubsection{Relevance Reward}
When evaluating the relevance between model-recommended entities and historical interaction context \( H \), this study draws upon the methodology of "LLM as a Judge"\cite{zheng2023judging} by employing a larger-scale model, DeepSeek-V3, as the judge. The relevance score calculation is divided into six steps:

\begin{enumerate}
    \item Retrieve relevant information \( \text{Song\_info}_i \) about the music entity \( M_i \) from the database.
    \item Drawing insights from \cite{dhole2025generativeproductrecommendationsimplicit}, we first employ the judge model to infer user needs \( N \) from \( H \), focusing on capturing implicit preferences beyond direct requests.   
    \item The judge assigns a score \( S_{1,i} \) (ranging from 0 to 1) indicating how well the music entity \( M_i \) satisfies the user's needs \( N \), based on the \( \text{Song\_info}_i \) to mitigate the impact of the judge's potential hallucinations.
    \item The judge evaluates whether the response text \( T_i \) addresses the user's needs \( N \) and assigns a score \( S_{2,i} \) (0 to 1).
    \item The judge rates the alignment between the response text \( T_i \) and the music entity \( M_i \) with a score \( S_{3,i} \) (0 to 1).
    \item The final Relevance Score is computed as: \(R_{\text{Relevance}}(H,M_i,T_i) = \lambda_1 S_{1,i} + \lambda_2  S_{2,i} + \lambda_3  S_{3,i}\ ,\ \sum_{i = 1}^3\lambda_i = 1\)
\end{enumerate}

\subsubsection{Personalization Reward}
We employ an online recommendation ranking model in WeChat Listen as the personalized reward model for evaluation. This model is trained on large-scale user interaction sequence data and can precisely quantify the semantic and preference-level alignment between candidate songs and users' historical behaviors. Consequently, it rewards recommendation results that exhibit continuity with users' established interests, yielding a personalization reward \(R_\text{Personalization}\).

\subsubsection{Repetition Penalty for Diversity}
To prevent the model from over-preferring certain popular songs and failing to learn the breadth of recommendation coverage, we employ a repetition penalty coefficient. We first obtain the factual reward and relevance reward corresponding to \(\overline{A}\), then group them by music entity \(M\). For a repeated recommendation group \(\overline{A}|_{M=m}\) of a specific entity \(m\), we sort the group internally based on \(R_{\text{Factuality}}(A) \times R_{\text{Relevance}}(H,A), A \in \overline{A}|_{M=m}\) to obtain the within-group rank \(Rank_{m}(A), A \in \overline{A}|_{M=m}\), and then progressively increase the penalty coefficient multiplier.
\begin{equation}
    \alpha_{\text{Repetition}}(A) = \alpha \times \left(Rank_{m}(A) - 1\right)
\end{equation}

\subsubsection{Hybrid Multi-verified Reward}
We construct a composite reward function that integrates the aforementioned four components, calculated as follows:
\begin{equation}
\label{eq:hybrid_reward}
    \begin{aligned}
R_{hybrid}(H,A_i)=&R_\text{Format}(A_i)\\&+(1-\alpha_\text{Repetition}(A_i))\times\mathbb I(R_\text{Format}(A_i)>0)\times R_\text{Factuality}(M_i)\\&\times[\texttt{Norm}(R_\text{Relevance}(H,M_i, T_i)+\texttt{Norm}(R_\text{Personalization}(H,M_i))]
\end{aligned}
\end{equation}
where $H$ is the history message and $A_i=(T_i, M_i)$ is the $i$-th generated sample in the rollout.

This study employs a multiplicative reward composition scheme to emphasize the tight interdependence between format adherence, factual accuracy, contextual relevance, and personalization. To promote output diversity, a repetition penalty is incorporated. Considering the potential conflicts and scale misalignment between relevance and personalization rewards, we apply a normalization function, \(\operatorname{Norm}(\cdot)\), to the set of reward values obtained from the candidate rollouts for each query. This function projects the raw rewards onto a common scale, enabling balanced reward integration. The policy model is subsequently optimized using GRPO \cite{shao2024deepseekmathpushinglimitsmathematical}.

\subsection{Self-Distillation: From Single-song Recommendation to List-wise Recommendation}
\hypertarget{sec:chapter_2_5}{}
\label{sec:self_dist}
To endow the model with list‑wise recommendation capability, i.e., the ability to recommend a set of songs as a coherent playlist in response to a user query, a straightforward approach is to construct a list‑wise QA dataset for fine‑tuning. However, we observe that in real‑world music recommendation systems, songs recommended to the same user often exhibit limited semantic relatedness, making it difficult to directly collect list‑wise data that reflects strong semantic cohesion.

To avoid the high cost of manual data collection, we design a self‑distillation algorithm that automatically converts single‑song recommendation data into list‑wise recommendation samples. The core idea is to perform repeated sampling with the WeMusic‑Base model, leveraging its intrinsic tendency to avoid repetitive recommendations, a property also utilized in the penalty mechanism described in Section~\ref{sec:rl_internal}. Concretely, we perform at most three rounds of resampling, each generating five candidate songs, until at least K distinct songs are collected (K=5 in experiments). After each resampling, the recommended songs are appended to the input message, i.e., the dialogue history, along with the same query, thereby preventing the model from recommending the same songs in subsequent rounds. The resulting list of songs is assigned to the \texttt{<music>} field for each original single‑song music search query, and only the playlist description in the \texttt{<text>} field is regenerated. For multi‑turn dialogues, we apply the same method in a sliding‑window fashion, synthesizing data round by round. Finally, we further conduct supervised fine‑tuning (SFT) on the synthesized multi‑turn list‑wise recommendation data using the WeMusic‑Base model. 

\section{WeMusic-Agent: Breaking Through the Boundary of Internalization}
\hypertarget{sec:chapter_3_wemusic_agent}{}
In this section, we introduce WeMusic‑Agent, a novel agentic framework designed for conversational music recommendation. Unlike existing music‑agent approaches that rely on external tools, which inherently suffer from inefficiency and limited personalization, we fully leverage the principles of \textbf{Knowledge Internalization} and \textbf{Agentic Tools} to construct a more capable music agent. Our framework extends the functional boundary of the agent while significantly enhancing its ability to deliver personalized user experiences.


\subsection{Internalized Model vs. Agent Model: Is There A Golden Rule in Music CRS?}
\hypertarget{sec:chapter_3_1}{}
\label{sec:golden_rule}
We begin by contrasting the strengths of Internal Music Models (e.g., WeMusic‑Base, WeMusic‑Base‑Dist) with those of existing Music Agent methods \cite{doh2025talkplay}. Internalized models offer a clear advantage in inference efficiency, as they generate responses and recommendations directly without requiring external tool calls. Moreover, while tool‑based agent approaches (e.g., Agentic RAG) often struggle to deliver personalized recommendations across diverse users, internalized models can better capture individual music preferences by encoding user‑song interaction signals. Nevertheless, knowledge internalization also presents limitations when handling queries that fall outside the model’s trained capability boundary. For out‑of‑distribution (OOD) questions, internalized models are prone to generating irrelevant or hallucinated content. This raises a fundamental question: is there a golden rule for music Conversational Recommendation Systems (CRS)?

\begin{figure*}[h]
\centering
\includegraphics[width=\textwidth]{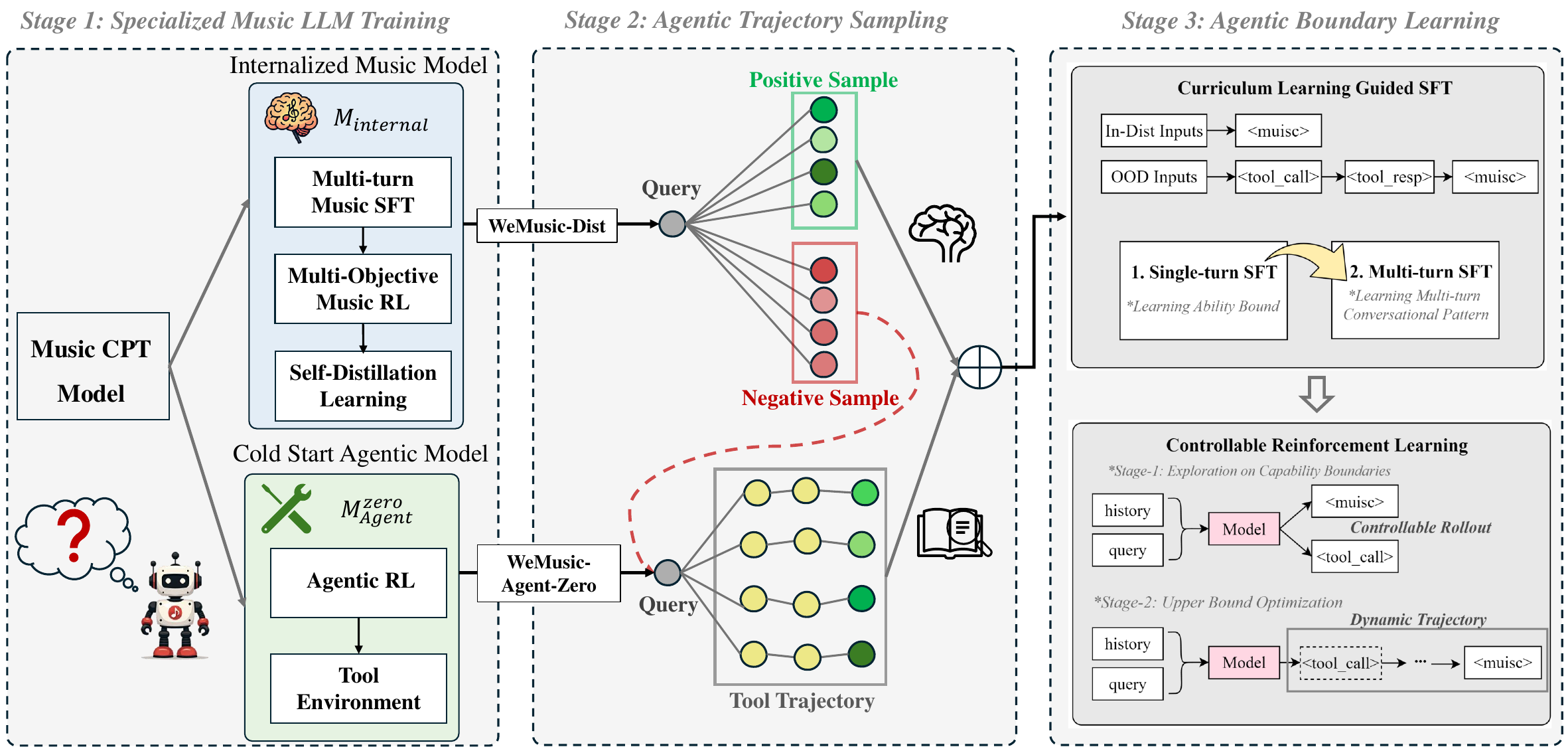} 
\caption{The Agent Boundary Learning framework of WeMusic-Agent}
\label{fig:agentic_bound_learning}
\end{figure*}

To address this challenge, we propose an Agent Boundary Learning framework, as illustrated in Fig.~\ref{fig:agentic_bound_learning}, and introduce WeMusic‑Agent, a unified model that integrates the respective strengths of knowledge‑internalized and agent‑based paradigms. Specifically, WeMusic‑Agent prioritizes the use of internalized knowledge for response generation and only call external tools for music search when necessary. Consequently, it retains the high inference efficiency and personalized song recommendation capabilities of internalized models, while also incorporating the more powerful capabilities of the agent‑based systems.

\subsection{Overview of WeMusic-Agent}
\hypertarget{sec:chapter_3_2}{}
WeMusic‑Agent is post‑trained on top of WeMusic via Agentic Boundary Learning. As depicted in Figure~\ref{fig:agentic_bound_learning}, Agentic Boundary Learning is designed to augment the capabilities of the internalized base model (WeMusic‑Base‑Dist) through efficient, on‑demand tool-utilization. The framework consists of three distinct stages.

\begin{enumerate}
    \item \textbf{Specialized Music LLM Training}. We train two independent specialized model respectively, including an internalized music LLM (WeMusic-Base-Dist, denoted as $M_{internal}$) and another agentic music LLM (WeMusic-Agent-Zero, denoted as $M_{agent}^{zero}$) that responses with music tool calling described in Section~\ref{sec:sec_agent_zero}.
    \item \textbf{Agentic Trajectory Sampling}. We collect real-world user queries from WeChat Listen, and generate samples with both $M_{internal}$ and $M_{agent}^{zero}$ to discover the capability boundaries of $M_{internal}$. Then we replace the negative samples from $M_{internal}$ with generations from $M_{agent}^{zero}$ for distillation.
    \item \textbf{Agentic Boundary Learning}. We combine positive samples from $M_{internal}$ and $M_{agent}^{zero}$ for Agentic Boundary Learning, including a curriculum learning guided SFT stage and a controllable reinforcement learning stage.
\end{enumerate}

\subsection{WeMusic-Agent-Zero: Music Agentic Model with Tool Calling for Cold Start}
\hypertarget{sec:chapter_3_3}{}
\label{sec:sec_agent_zero}
To train an agentic model capable of deciding when to call external music‑search tools, a straightforward approach is to construct a dataset containing both agentic (out‑of‑knowledge) and non‑agentic (in‑knowledge) samples. However, we observe that existing general models typically perform poorly on tasks involving private music tool calling (APIs of precise music search, fuzzy music search and general web search) and personalized music ranking. 

To accelerate cold start, we train a specialized agentic model, WeMusic‑Agent‑Zero $M_{Agent}^{Zero}$, based on the CPT model. We first design a structured prompt that specifies the available music search tools and their calling protocols. Agentic reinforcement learning is then performed using GRPO. In this setup, we simulate an interactive environment that supports function calling during training and evaluates the final response of each sampled trajectory. The reward function incorporates the same format, relevance, and personalization terms defined in Section~\ref{sec:rl_internal}, and introduces an additional diversity reward, measured as the number of distinct valid songs in a recommendation, to encourage list‑wise recommendation quality.

In this manner, we prepare two specialized model for list-wise music recommendation, i.e., an internalized model $M_{internal}$ and an agentic model with tool calling $M_{agent}^{zero}$. As discussed in Section~\ref{sec:golden_rule}, $M_{internal}$ has higher efficiency and stronger personalization for user preference, and $M_{agent}^{zero}$ has broader scope of capabilities for Out-Of-Knowledge queries. Then we use $M_{internal}$ as the anchor model to define capability boundary (In-Knowledge or Out-Of-Knowledge) and use samples generated from $M_{agent}^{zero}$ as cold start data for learning necessary tool calls.

\subsection{Agentic Trajectory Sampling: Discovering the Boundary of Internalization}
\hypertarget{sec:chapter_3_4}{}
To delineate the capability boundary of internalized knowledge for $M_{internal}$, we first construct a sampling dataset. In particular, we collect 50K real‑world user queries from WeChat Listen and categorize them into two types based on the scope of internalized knowledge during CPT(see Section~\ref{sec:cpt}):
\begin{enumerate}
    \item In-Distribution Samples: queries that can be well solved by the internalized model $M_{internal}$, especially for queries of music included in our pretraining corpus.
    \item Out-Of-Distribution Samples: queries that can not be solved by internalized model $M_{internal}$, such as queries of music not included in the pretraining corpus or music related to the latest/hottest events.  
\end{enumerate}
We then perform trajectory sampling on the training set to identify the agentic capability boundary. Specifically, for each training sample, we obtain rollouts from $M_{internal}$ and evaluate them using the relevance reward function defined in Section~\ref{sec:rl_internal}. Based on the relevance scores, the generated samples are divided into two categories: positive samples ($Pos_{internal}$), where at least one recommended song achieves a relevance score of 6 or higher (as defined in Section~\ref{sec:rl_internal})), and negative samples ($Neg_{internal}$), which fail to meet this threshold. For each negative sample, we further generate a corresponding trajectory that includes function callings using WeMusic-Agent-Zero; these are denoted as $Pos_{agent}$. In this formulation, $Pos_{internal}$ corresponds to the comfort zone of $M_{internal}$, whereas $Neg_{internal}$ reflects its out-of-knowledge zone.

\subsection{WeMusic-Agent-M1: Efficient Agentic Boundary Learning}
\hypertarget{sec:chapter_3_5}{}
\label{sec:agentic_boundary_learning}
To learn agentic boundary (when to call external tools and the planning for tool choices), we conduct supervised finetuning with curriculum learning with single-turn samples in stage 1 and multi-turn samples in stage 2. Specifically, we use 50k single-turn samples in stage 1, and use 10k multi-turn samples, each having 5$\sim$10 turns, in stage 2. The single-turn data is sampled from the multi-turn samples in stage 2. In our experiments, we find that curriculum learning in a step-by-step manner, from single-turn to multi-turn, can help the model learn the capability boundary better. In the first stage, the model trained on single-turn data reached 93\% accuracy of tool calling (when to use tools) on the test set. And we surprisingly find that the model retains comparable tool calling accuracy during the second stage on multi-turn dialogue scenarios.

To differentiate between the output of single-song recommendations and list-of-songs recommendations, we use $\overline{A}=\{A_i\}_L=\{(T_i,M_i)\}_L$ to denote single-song output ($M_i$ is a single song) and use $\overline{\mathcal{A}}=\{\mathcal{A}_i\}_L=\left\{(T_i,\{M_{i,j}\}_{j=1}^K)\right\}_L$ to denote list-of-song output, where $\{M_{i,j}\}_{j=1}^K$ is a set of songs and $K$ is the number of output songs in $\mathcal{A}_i$. Finally, we design a controllable reinforcement learning algorithm to improve the tool call efficiency of WeMusic further, including the \textbf{Capability Boundary Exploration} and the \textbf{Upper Bound Optimization}.

To improve the exploration efficiency of Agentic RL for ability boundary learning, we design the Controllable Boundary Exploration algorithm, including a controllable rollout process and an agentic hybrid reward function. In each training step, we conduct controllable rollout to ensure that both the internalized (non-agentic) response and agentic response are observable given the same prompt. Specifically, we generate half \textbf{agentic samples} $\overline{\mathcal{A}}^{\ agentic}$ with tool calls and half \textbf{internalized samples} $\overline{\mathcal{A}}^{\ internal}$ with knowledge internalization for each rollout by controlling the prefix special tokens (\texttt{<intention>} or \texttt{<tool\_call>}). Specifically, we sample four internalized outputs and four agentic trajectory outputs, a total of eight samples for each training sample in our experiments. Then we reuse the reward function defined in Section~\ref{sec:rl_internal}. Different from WeMusic-Base which recommend a single song for each generation, the output of WeMusic-Agent is a list of songs, inherited from WeMusic-Base-Dist (see Section~\ref{sec:self_dist}) and WeMusic-Agent-Zero (see Section~\ref{sec:sec_agent_zero}). We thus transfer the hybrid reward function of single-song-recommendation $R_{hybrid}$ defined in Eq.~\ref{eq:hybrid_reward} into the list-wise hybrid reward function:

\begin{equation}
\label{eq:hybrid_reward}
    \begin{aligned}
 R_{hybrid}^{list}(H, \mathcal{A}_i) = &R_\text{Format}(\mathcal{A}_i)\\& + \frac{K}{N_{max}} \times \sum_{j=1}^K \Big{\{} \mathbb{I}(R_\text{Format}(M_{i,j})>0)\times R_\text{Factuality}(M_{i,j})\\&\times[\texttt{Norm}(R_\text{Relevance}(H,M_{i,j}, T_i)+\texttt{Norm}(R_\text{Personalization}(H,M_{i,j}))] \Big{\}},
\end{aligned}
\end{equation}
where $H$ is the history message, $\mathcal{A}_i=(T_i,\{M_{i,j}\}_{j=1}^K)$ is the $i$-th generated sample in the rollout, and $N_{max}$ is a hyper-parameter controlling the largest number of possible recommended songs. And we set $N_{max}=5$ in our experiments.

To improve the overall efficiency, we add a multiplier as \textbf{discount factor} (denoted as $\gamma$) to rewards of agentic samples. With the discount factor $\gamma$, we encourage the model to make the most of internalized knowledge and call external searching tools when necessary, and we set $\gamma=0.8$ in our experiments. The agentic hybrid reward function $R_{controllable}$ is defined as follows:

\begin{equation}
\label{eq:agentic_hybrid_reward}
    R_{hybrid}^{agentic} = \left\{ 
	\begin{aligned}
	&R_{hybrid}^{list}(H, \mathcal{A}_i)  &  &\text{if}\ \mathcal{A}_i \in\ \overline{\mathcal{A}}^{\ internal}\\
	&\gamma \cdot R_{hybrid}^{list}(H, \mathcal{A}_i) & &\text{else if}\ \mathcal{A}_i \in\ \overline{\mathcal{A}}^{\ agentic}\\
	\end{aligned}
	\right.
\end{equation}
where $H$ denotes the conversational history, $A_i$ denotes the $i$-th response in the rollout. Then we optimize the policy model with GRPO~\cite{shao2024deepseekmathpushinglimitsmathematical} algorithm. The main idea of controllable boundary exploration is to learn when to call music searching tools, we further boost the music recommendation perfromance and tool utilization efficiency via Upper Bound Optimization. In this stage, we lift the restrictions in the previous controllable rollout process and re-empowering the policy model to decide when to invoke tools. And we directly optimize the model with the agentic hybrid reward function defined in Eq.~\ref{eq:agentic_hybrid_reward}.

\section{WeMusic-Bench: Benchmark and Metrics for Music Recommendation}
\hypertarget{sec:chapter_4_benchmark}{}
\label{sec:chapter_4_benchmark}






We proposed a benchmark for conversational music recommendation named WeMusic-Bench by collecting evaluation queries from two realistic sources where some examples are out-of-knowledge boundary for our internalized model. One source is extracted from the logs in online WeChat Search scenario and the other is from the records of our internal experience users while testing our model. For the online WeChat Search logs, we clean and reformulate the top search keywords or short sentences into question form. For the experience records of our model, we annotate song search queries from a variety of conversational sessions. To enrich the query complexity, original simple queries are augmented by adding more search restrictions. We split the dataset into three-level complexity according to the number of restrictions in the query: \textbf{Level-1} with one restriction, \textbf{Level-2} with two restrictions, and \textbf{Level-3} with over two restrictions. Finally, there are about 1.3k testing samples covering diverse aspects such as genre, mood, language, region, style, period, rhythm, tempo, instrument, and etc. 

To systematically evaluate the model's effectiveness, this study constructs a three-dimensional evaluation framework (Relevance, Diversity, Personalization) for quantitative analysis of Music Conversational Recommendation Model (Music CRM) \(Model\), evaluated on our test set \(\overline{D}=\{H_i\}, i\in[1,L]\), where \(L\) indicates the size of test set.

\subsubsection{Relevance/Personalization Evaluation} We employ the same evaluation methodology used for reward generation during the training process in Section~\ref{sec:rl_internal} to obtain the assessment scores.

\subsubsection{Diversity Evaluation}
To measure the breadth of coverage for music entities \( M \) recommended by the model, we adopt the concept of "effective diversity"\cite{shypula2025evaluatingdiversityqualityllm}. For each dialogue history \( H_i \) in the test set, the model generates a set of \( K \) recommendations \( \overline{A}_i = \{A_{i,1}, A_{i,2}, ..., A_{i,K}\} \) (where \( K \) is a hyperparameter set to 5 in our experiments). The model's final diversity score is determined by calculating the proportion of all valid and distinct recommendation pairs in this set. Specifically:

\textbf{Effectiveness Criterion} \(\mathrm{Effective(A)}\): A response \(A\) is considered valid if and only if its format is correct, the Factuality check passes, and \(R_\text{Relevance} > 0.6\). This ensures that recommendations considered for diversity must first satisfy basic criteria for factuality and relevance.

\textbf{Semantic Difference} (\( d_{\mathrm{sem}}(A_i, A_j) \)): For any two distinct recommendations \( A_i \) and \( A_j \), their semantic difference is defined as:
    \[
    d_{\mathrm{sem}}(A_i, A_j) = \begin{cases}
    1 & \begin{aligned} \text{if } &\mathrm{Effective}(A_i) \text{ and } \mathrm{Effective}(A_j) \\ & \text{ and } M_i \ne M_j \end{aligned}\\
    0 & \text{otherwise}
    \end{cases}
    \]
    Only when both recommendations are valid and their recommended music entities \( M_i \) and \( M_j \) differ do they contribute to diversity (score of 1).

\textbf{Diversity Score Calculation}: For the song-list recommendation task, we just use the number of valid recommended songs in each completion as its diversity score. Then for the single-song recommendation task, we propose a novel diversity metric to evaluate the model with multiple completions for each query as follows: given the dialogue history \( H_i \) and the semantic scores \( d_{\mathrm{sem}} \) across all possible recommendation pairs in \( \overline{A}_i \) (totaling \( \binom{K}{2} \) pairs), the model's overall diversity score is the mean of diversity scores across all dialogue histories \( H_i \) where \(L\) denotes the size of the test set and \(K\) represents the number of generations for each \(H\):
    \[
    \mathrm{Diversity}(Model) = \frac{1}{L} \sum_{i=1}^{L} \left( \frac{1}{\binom{K}{2}} \sum_{1 \le p < q \le K} d_{\mathrm{sem}}(A_{i,p}, A_{i,q}) \right)
    \]

\section{Experiments and Discussion}
We evaluate the performance of WeMusic series models on WeMusic-Bench in this section.
\hypertarget{sec:chapter_5_exp}{}
\label{sec:chapter_5_exp}
\begin{figure*}[h]
\centering
\includegraphics[width=\textwidth]{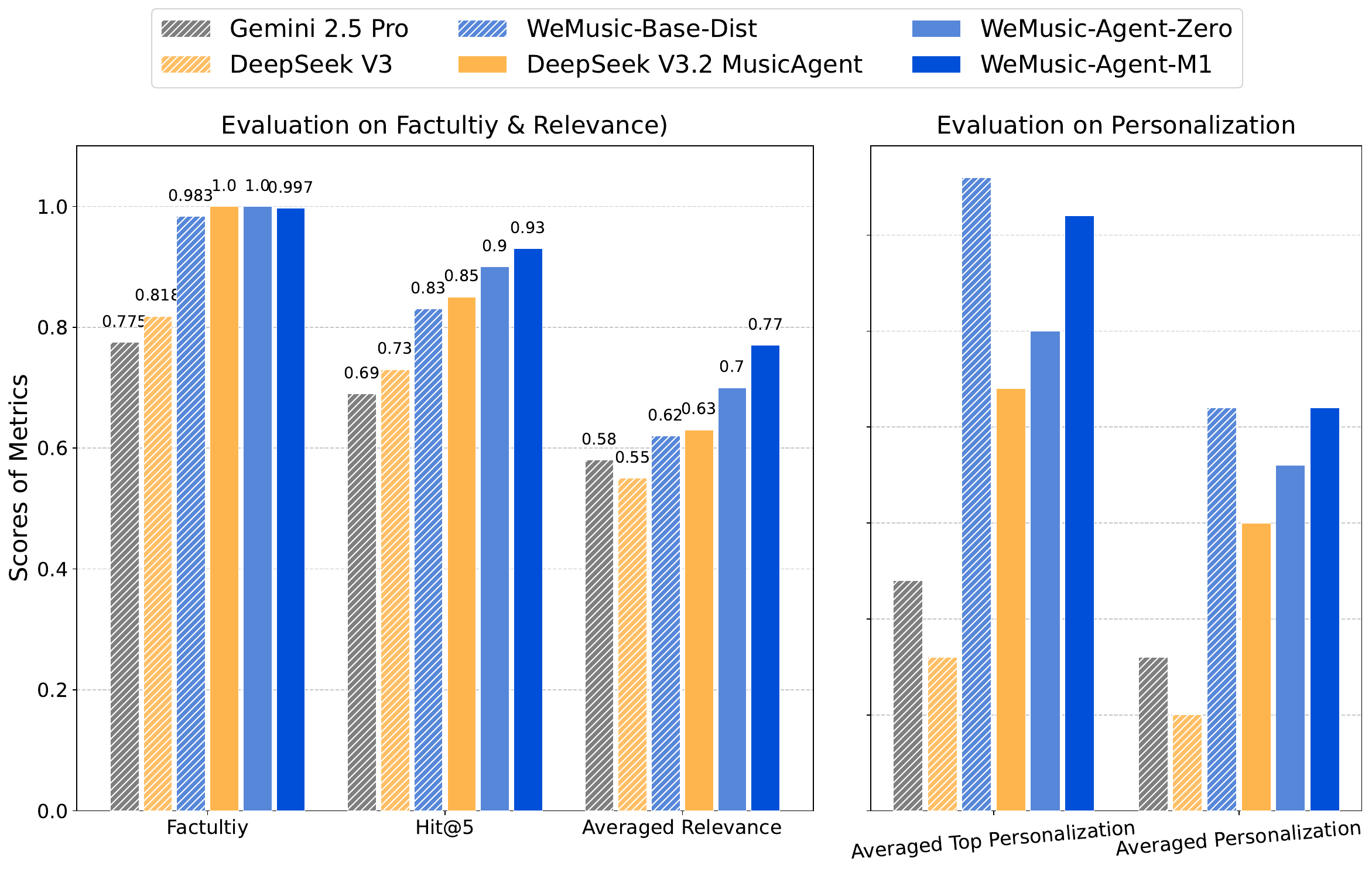} 
\caption{Comparison of WeMusic and other state-of-the-art LLMs on WeMusic-Bench. And we evaluate WeMusic-Base-Dist and WeMusic-Agent in the size of 32B parameters.}
\label{fig:main_res}
\end{figure*}

\subsection{Main Results}
\hypertarget{sec:chapter_5_1}{}
We compare the performance of WeMusic series models and other open-sourced or closed-sourced models in this section, including Gemini-2.5-pro, DeepSeek-V3 and DeepSeek-V3.2-MusicAgent. Specifically, we use the same prompts for WeMusic-Base-Dist, Gemini-2.5-pro and DeepSeek-V3 for conversational music recommendation with internalized knowledge only. And we use the same prompts for WeMusic-Agent-Zero and DeepSeek V3.2 MusicAgent for agentic conversational music recommendation with the same set of external music tools.
The evaluation results of each model on the personalized music recommendation benchmark are shown in Figure \ref{fig:main_res}. The experimental results reveal a clear performance hierarchy: specialized music models significantly outperform general-purpose models (Gemini 2.5 Pro and DeepSeek-V3) across all evaluation metrics on WeMusic-Bench. This strongly validates the necessity of domain-specific specialization in model design.

Among the specialized models, WeMusic-Agent-M1 demonstrates comprehensive performance superiority. It ranks first in three core metrics: recommendation recall ('Hit@5', 0.93), average relevance ('Averaged Relevance', 0.77), and average personalization score ('Averaged Personalization'), while maintaining near-perfect factual probability. This establishes its leading position as the most robust and comprehensively capable personalized music recommendation model currently available.

In-depth analysis indicates that the agentic ability is crucial for enhancing the model's fundamental recommendation capability. By comparing models within the WeMusic series, it is evident that the agent-enabled WeMusic-Agent-Zero and WeMusic-Agent-M1 consistently surpass the non-agent baseline model WeMusic-Base-Dist in both 'Hit@5' and 'Averaged Relevance' metrics. This is because there are Out-of-Knowledge tasks in WeMusic-Bench according to the internalized knowledge of WeMusic-Base. This suggests that the tool invocation and planning mechanisms introduced by the agent effectively expand the model's capability boundaries, thereby systematically improving the accuracy and overall quality of the recommendations.

Regarding the key dimension measuring personalization depth, the experimental results reveal two distinct technical pathways, each with its own advantages. WeMusic-Base-Dist excels at capturing users' most prominent preference features, achieving the best 'Average Top Personalization'. In contrast, WeMusic-Agent-M1 reaches a comparable top-tier level in the more comprehensive 'Averaged Personalization', while demonstrating significantly stronger fundamental recommendation and interaction capabilities. This highlights the superiority of the WeMusic-Agent-M1 in understanding and aligning with complex user needs.

\subsection{Discussions on the Effects of MusicCPT}
\hypertarget{sec:chapter_5_2}{}
\begin{figure}[htbp]
    \centering
    \begin{subfigure}[b]{0.43\textwidth}
        \includegraphics[width=\linewidth]{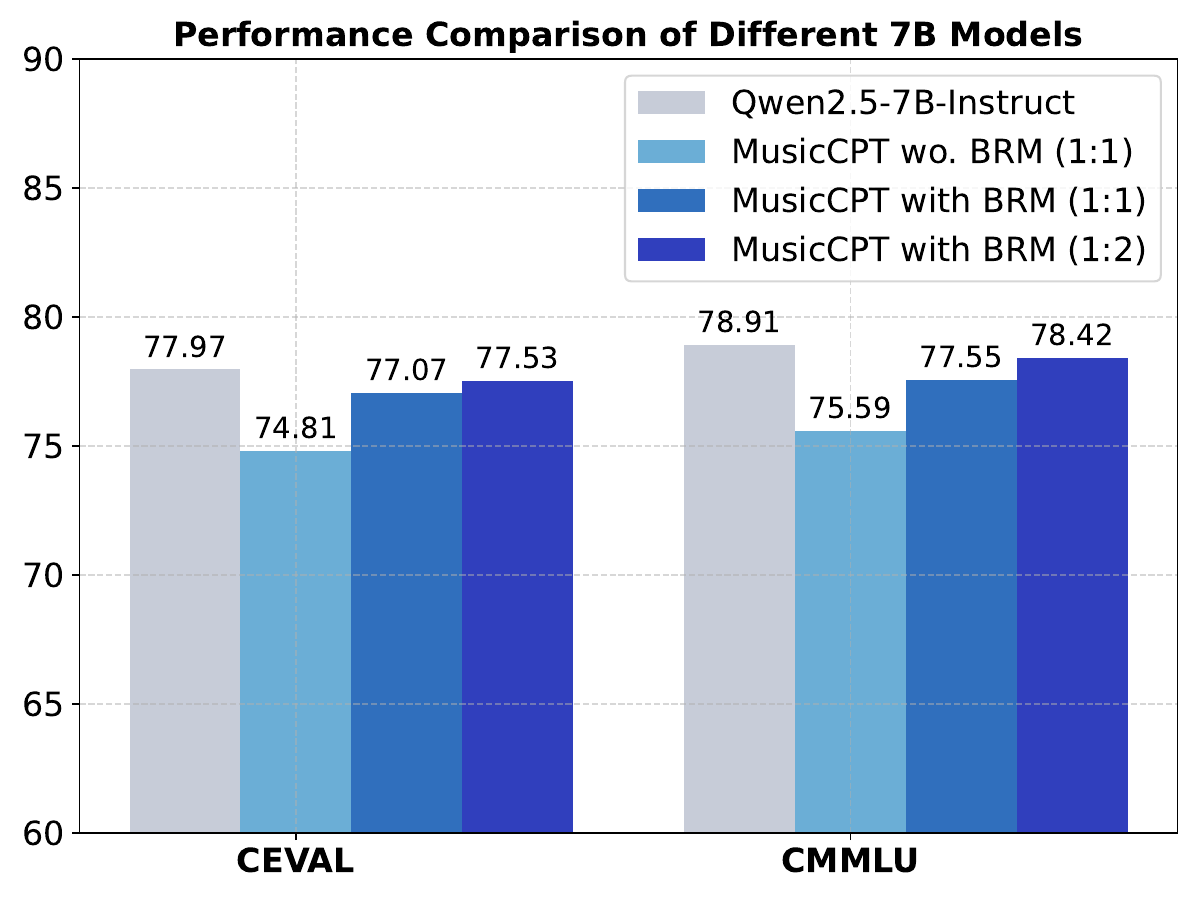}
        \caption{Performance comparison of 7B model}
        \label{fig:sub1}
    \end{subfigure}
    \hspace{10pt}
    \begin{subfigure}[b]{0.43\textwidth}
        \includegraphics[width=\linewidth]{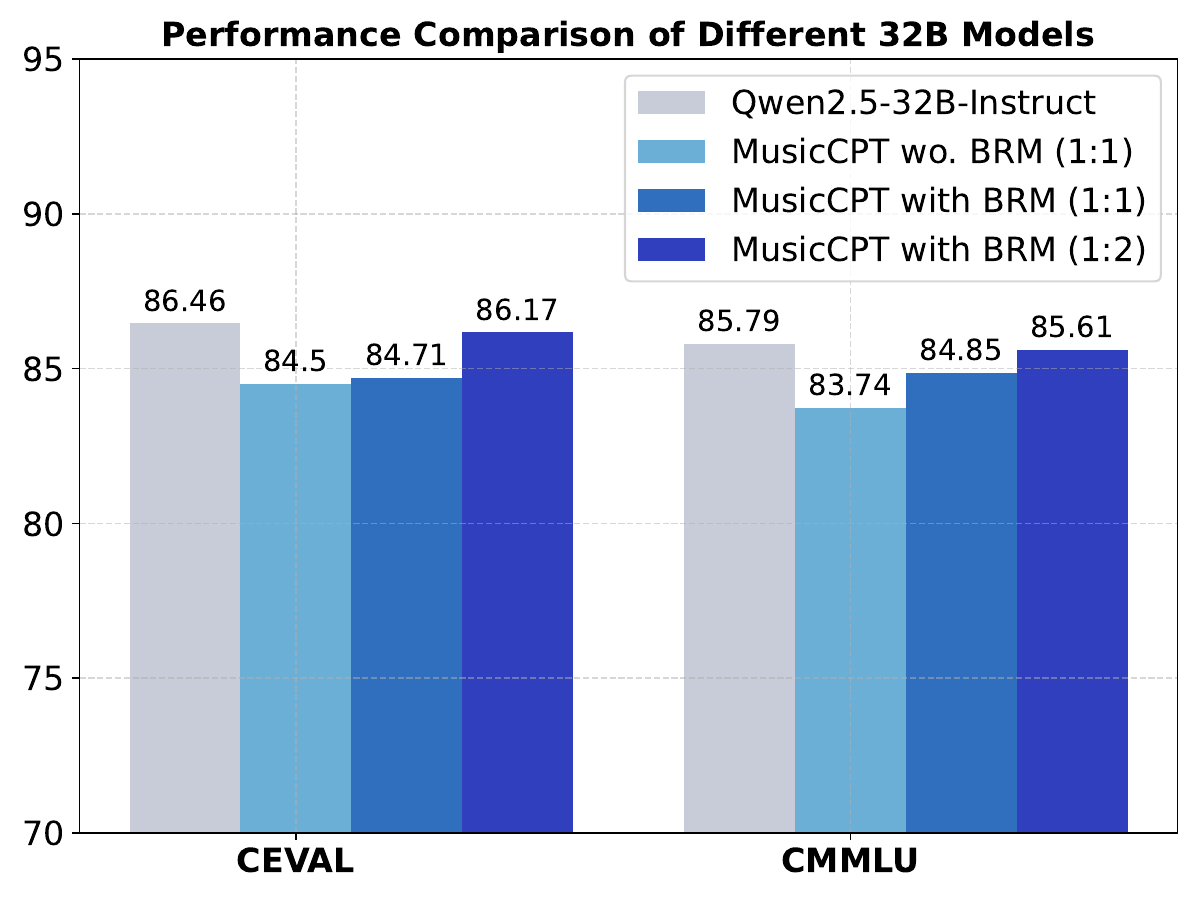} 
        \caption{Performance comparison of 32B model}
        \label{fig:general_result_bar}
    \end{subfigure}
    \caption{Evaluation on CEVAL and CMMLU benchmarks to show the effectiveness of our BRM to  preserve general knowledge. "wo. BRM" and "with BRM" mean that the model trained without and with base reference model constraint respectively. "(1:1)" and "(1:2)" represent the ratio between music and general domain.}
    \label{fig:both}
\end{figure}


To understand the effect of our MusicCPT continual pretraining strategy, we first build a factual QA dev set MusicSimpleQA following the construction recipe of MuCPT~\cite{tian2025mucpt} and the OpenAI SimpleQA ~\cite{wei2024measuring} benchmark for factual knowledge probing.  Each example is designed to have a short and unique answer that is stable over time, so that the evaluation mainly reflects
how well the model memorizes music-related entities and attributes rather than its reasoning or dialogue skills. To endow our CPT model with music domain knowledge and weak instruction-following capabilities to evaluate on simple QA benchmarks, we follow \cite{demify_cpt} to continue training in instruct series checkpionts \cite{Yang2024Qwen25TR} with both document and instruction data.

\begin{table}[h]
\centering
\caption{Ablation Study of Token-Soft-Scoring on CPT}
\begin{tabular}{lc}
\hline
Method             & MusicSimpleQA \cite{tian2025mucpt}   \\ 
\hline
Qwen-7B-NextTokenPrediction   & 0.5499          \\
Qwen-7B-RHO-1      & 0.5699          \\
Qwen-7B-Token-Soft-Scoring (ours)      & \textbf{0.6119} \\ \hline
\hline
Qwen-32B-NextTokenPrediction   & 0.6253          \\
Qwen-32B-RHO-1      & 0.6472          \\
Qwen-32B-Token-Soft-Scoring (ours)      & \textbf{0.6731} \\ \hline
\end{tabular}
\label{tab:eval_token_soft_scoring}
\end{table}

On MusicSimpleQA we compare three training paradigms with the same data configuration in stage 1 of CPT: 
(i) standard next-token prediction (NTP), 
(ii) the token-filtering based RHO-1 \cite{lin2024rho} objective, and
(iii) our token-soft-scoring MusicCPT. 
As shown in Table~\ref{tab:eval_token_soft_scoring}, both RHO-1 and MusicCPT significantly outperform the NTP baseline on 7B and 32B models, which confirms that the raw corpus indeed contains a non-trivial amount of noisy segments and that treating all tokens equally hurts the learning of salient signals.  Moreover, MusicCPT consistently achieves the best accuracy across all model sizes.  Unlike RHO-1, which hard filters low scoring tokens, our token-soft-scoring assigns fine-grained weights within the same sentence and reweights the loss accordingly.  This allows the model to focus more on high quality fragments while still preserving the original context, thereby avoiding the semantic discontinuity that may be introduced by aggressive tokendropping.  This property is particularly important for music related descriptions that often rely on rich contextual cues.

Since WeMusic-Base is primarily deployed in the WeChat ecosystem for Chinese conversational music scenarios, we further examine the impact of MusicCPT on general Chinese knowledge capabilities using the widely-used  CEVAL\cite{nguyen2024ceval} and CMMLU\cite{li2024cmmlu} benchmarks. 
Figure~\ref{fig:both} reports the results for 7B and 32B models under different CPT configurations.   Compared with models without BRM, after introducing BRM self-distillation, the 7B model obtains a substantial improvement on both benchmarks, while the 32B model, whose baseline is already strong, still enjoys up to about 1\% absolute gain. Under a music:general data ratio of $1{:}2$, the CPT+BRM models perform on par with the original CPT initialization on CEVAL and CMMLU, showing that large-scale music specialization can be achieved without sacrificing general knowledge.
We also meticulously investigate how the mixture ratio between music domain and general domain data influences model performance on small proxy models in appendix~\ref{mixture_section}.

\begin{table}[t]
    \centering
    \setlength\tabcolsep{10pt}
    \begin{tabular}{cccc}
    
    \toprule
    \multirow{2}{*}{Model} & \multicolumn{3}{c}{Temperature}  \\
      &0.5&0.7& 1.0 \\
    \midrule
    Gemini-2.5-pro & 0.179\scalebox{0.8}{±0.001} & 0.219\scalebox{0.8}{±0.003} & 0.233\scalebox{0.8}{±0.005} \\
    DeepSeek V3 & 0.218\scalebox{0.8}{±0.006} & 0.269\scalebox{0.8}{±0.004} & 0.307\scalebox{0.8}{±0.008} \\
    Qwen3-235B-A22B-Instruct & 0.217\scalebox{0.8}{±0.003} & 0.261\scalebox{0.8}{±0.006} & 0.261\scalebox{0.8}{±0.006} \\
    Qwen2.5-32B-Instruct & 0.084\scalebox{0.8}{±0.002} & 0.093\scalebox{0.8}{±0.002} & 0.089\scalebox{0.8}{±0.001} \\
    Qwen2.5-32B-MusicCPT & 0.160\scalebox{0.8}{±0.002} & 0.280\scalebox{0.8}{±0.004} & 0.300\scalebox{0.8}{±0.003} \\
    WeMusic-Base-32B & \textbf{0.525}\scalebox{0.8}{±0.003} & \textbf{0.598}\scalebox{0.8}{±0.003} & \textbf{0.667}\scalebox{0.8}{±0.004} \\
    \bottomrule
\end{tabular}
\caption{Comparison of Diversity Scores on WeMusic-Bench for LLMs with internalized knowledge only.}
\label{tab:diversity_eval}
\end{table}

\subsection{Effectiveness of Diversity Rewards in WeMusic-Base}
\hypertarget{sec:chapter_5_4}{}
In this section, we discuss the impact of recommendation diversity for the internalized model (WeMusic-Base), because it is critical for the capability boundary of internalized models and is the foundation of the WeMusic-Agent model. Specifically, we conduct empirical studies to compare the diversity of conversational music recommendations between WeMusic-Base and other SOTA LLMs in this section. To evaluate the diversity for models of single-song recommendations, we use the diversity score introduced in the section~\ref{sec:chapter_4_benchmark} as the evaluation metric.  As shown in the table~\ref{tab:diversity_eval}, we compare the diversity of different LLMs that reply with internalized knowledge only with varing settings of temperature. And the result show that our WeMusic-Base-32B model improves the recommendation diversity of conversational music recommendation tasks by more than double compared to our MusicCPT model and other state-of-the-art LLMs, which demonstrates the effectiveness of the diversity reward, i.e., the repetition penalty introduced in the section~\ref{sec:rl_internal}.





\begin{figure*}[h]
\centering
\includegraphics[width=0.8\textwidth]{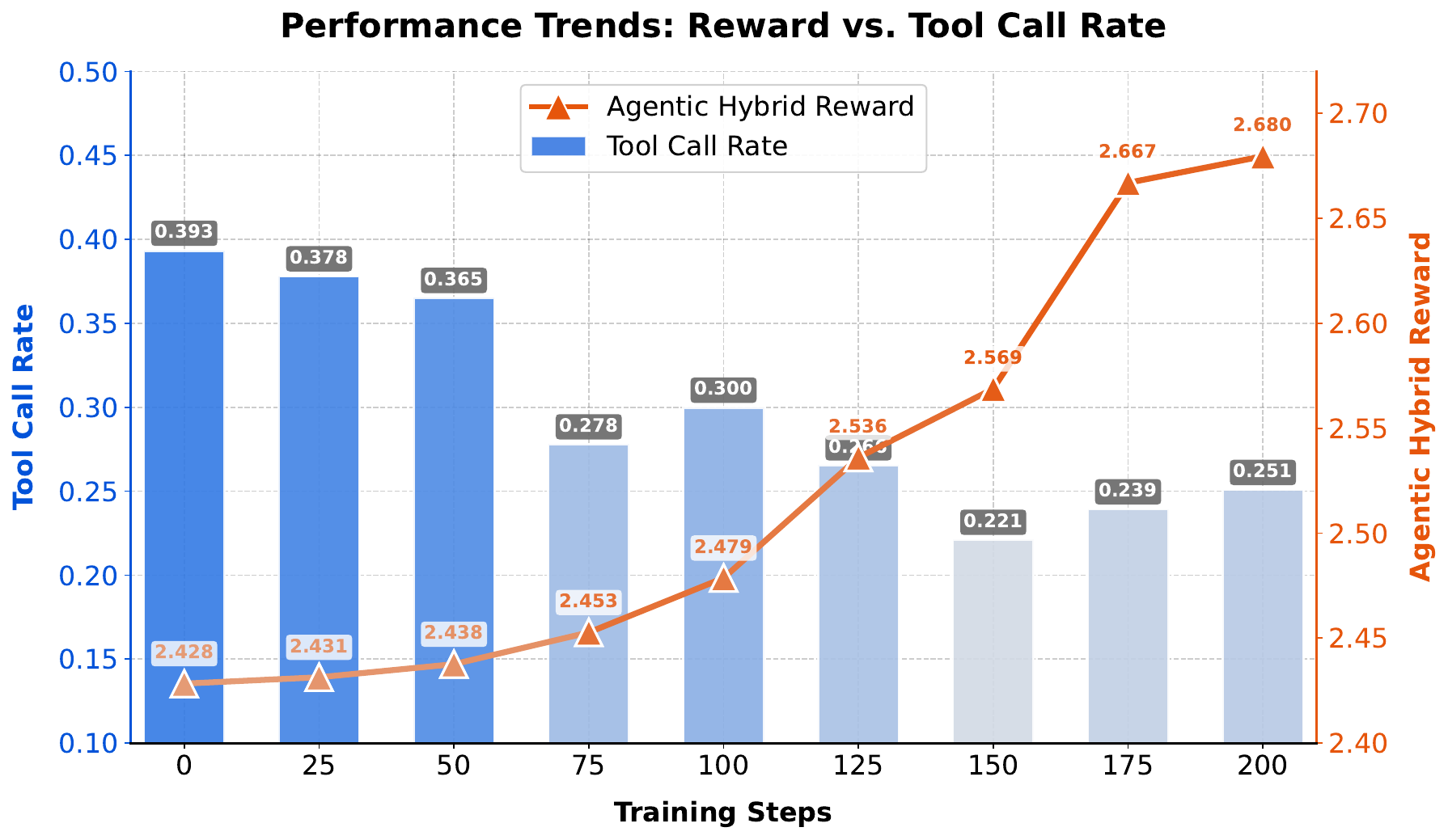} 
\caption{The trends of agentic hybrid rewards and tool call rate during the controllable reinforcement learning process of WeMusic-Agent-M1}
\label{fig:reward_tool_rate}
\end{figure*}

\subsection{Discussions on Rewards and Tool Efficiency of WeMusic-Agent}
\hypertarget{sec:chapter_5_3}{}
As introduced in the section~\ref{sec:agentic_boundary_learning}, the motivation of controllable reinforcement learning is to discover the capability boundary (when to use internal knowledge and when to use external tools), and control the tool calling rate to improve the overall response efficiency. We conduct empirical study on this. As shown in the Figure~\ref{fig:reward_tool_rate}, we validate the trend of the hybrid reward and tool calling rate during the controllable reinforcement learning  process of WeMusic-Agent-M1. We observe that as training progresses, the proportion of tool usage is gradually decreasing to a stable level ($\sim25\%$ on the test set), while the overall hybrid reward of WeMusic-Agent-M1 increases. This phenomenon indicates that our method helps the model generate more satisfactory song recommendations while balancing the usages of searching tools and the internalized knowledge to some extent.

\section{Conclusion}
\hypertarget{sec:chapter_6_conclusion}{}
\label{sec:chapter_6_conclusion}
In this paper, we propose a training framework for LLM-based Music Conversational Recommendation (CRS). We systematically discuss the advantages and disadvantages of the knowledge internalization methods and the agentic methods. Under this framework, we first propose WeMusic-Base trained in a multi-stage manner from Qwen series models, including continual pretraining on music corpus at scale and post-training for single-song recommendation tasks. Then we conduct self-distillation on WeMusic-Base to get WeMusic-Base-Dist for song-list recommendation tasks. To solve the explicit capability boundary of internalization model, we further propose WeMusic-Agent, trained with agentic boundary learning and supported for necessary agentic tools calling for out-of-knowledge music recommendations. Besides, we also propose a benchmark for evaluation of CRS, which is collected real-world data from WeChat Listen. Ultimately, WeMusic-Agent-M1 achieves performance outperformed to existing open-sourced and closed-sourced models in CRS tasks.



\section{Disclaimer and Security Announcement}
\hypertarget{sec:chapter_7_disclaimer}{}
\label{sec:chapter_7_disclaimer}
This project is exclusively established for academic purposes, aiming to facilitate communication and learning. We aim to propose a training and evaluation framework of conversational recommendation in the music area. We did not use any non-anonymized user privacy data within the training framework. If you use our project for commercialization or product development, please carefully review the legality and compliance of the training data and application scenarios. Please ensure that you address any authorization issues related to the dataset. You bear full responsibility for any problems arising from the usage of unauthorized datasets for training, as well as any resulting consequences.


%
%
%
\hypertarget{sec:reference}{}
\bibliographystyle{splncs04}
\bibliography{ref}

@String{Computing = "Computing" }

@article{shypula2025evaluatingdiversityqualityllm,
  title={Evaluating the diversity and quality of llm generated content},
  author={Shypula, Alexander and Li, Shuo and Zhang, Botong and Padmakumar, Vishakh and Yin, Kayo and Bastani, Osbert},
  journal={arXiv preprint arXiv:2504.12522},
  year={2025}
}

@article{Jannach_2021,
  title={A survey on conversational recommender systems},
  author={Jannach, Dietmar and Manzoor, Ahtsham and Cai, Wanling and Chen, Li},
  journal={ACM Computing Surveys (CSUR)},
  volume={54},
  number={5},
  pages={1--36},
  year={2021},
  publisher={ACM New York, NY, USA}
}

@article{palumbo2025text2trackspromptbasedmusicrecommendation,
  title={Text2Tracks: Prompt-based Music Recommendation via Generative Retrieval},
  author={Palumbo, Enrico and Penha, Gustavo and Damianou, Andreas and Garc{\'\i}a, Jos{\'e} Luis Redondo and Heath, Timothy Christopher and Wang, Alice and Bouchard, Hugues and Lalmas, Mounia},
  journal={arXiv preprint arXiv:2503.24193},
  year={2025}
}

@article{shao2024deepseekmathpushinglimitsmathematical,
  title={Deepseekmath: Pushing the limits of mathematical reasoning in open language models},
  author={Shao, Zhihong and Wang, Peiyi and Zhu, Qihao and Xu, Runxin and Song, Junxiao and Bi, Xiao and Zhang, Haowei and Zhang, Mingchuan and Li, YK and Wu, Yang and others},
  journal={arXiv preprint arXiv:2402.03300},
  year={2024}
}

@article{deepseek_v3,
  title={Deepseek-v3 technical report},
  author={Liu, Aixin and Feng, Bei and Xue, Bing and Wang, Bingxuan and Wu, Bochao and Lu, Chengda and Zhao, Chenggang and Deng, Chengqi and Zhang, Chenyu and Ruan, Chong and others},
  journal={arXiv preprint arXiv:2412.19437},
  year={2024}
}

@article{dhole2025generativeproductrecommendationsimplicit,
  title={Generative Product Recommendations for Implicit Superlative Queries},
  author={Dhole, Kaustubh D and Vedula, Nikhita and Kuzi, Saar and Castellucci, Giuseppe and Agichtein, Eugene and Malmasi, Shervin},
  journal={arXiv preprint arXiv:2504.18748},
  year={2025}
}

@article{qwen2,
  title={Qwen2 technical report},
  author={Team, Qwen},
  journal={arXiv preprint arXiv:2407.10671},
  year={2024}
}

@article{qwen3,
  title={Qwen3 technical report},
  author={Yang, An and Li, Anfeng and Yang, Baosong and Zhang, Beichen and Hui, Binyuan and Zheng, Bo and Yu, Bowen and Gao, Chang and Huang, Chengen and Lv, Chenxu and others},
  journal={arXiv preprint arXiv:2505.09388},
  year={2025}
}

@article{zheng2023judging,
  title={Judging llm-as-a-judge with mt-bench and chatbot arena},
  author={Zheng, Lianmin and Chiang, Wei-Lin and Sheng, Ying and Zhuang, Siyuan and Wu, Zhanghao and Zhuang, Yonghao and Lin, Zi and Li, Zhuohan and Li, Dacheng and Xing, Eric and others},
  journal={Advances in neural information processing systems},
  volume={36},
  pages={46595--46623},
  year={2023}
}

@article{afchar2022explainability,
  title={Explainability in music recommender systems},
  author={Afchar, Darius and Melchiorre, Alessandro and Schedl, Markus and Hennequin, Romain and Epure, Elena and Moussallam, Manuel},
  journal={AI Magazine},
  volume={43},
  number={2},
  pages={190--208},
  year={2022}
}

@inproceedings{sun2018conversational,
  title={Conversational recommender system},
  author={Sun, Yueming and Zhang, Yi},
  booktitle={The 41st international acm sigir conference on research \& development in information retrieval},
  pages={235--244},
  year={2018}
}

@article{doh2025talkplay,
  title={TalkPlay-Tools: Conversational Music Recommendation with LLM Tool Calling},
  author={Doh, Seungheon and Choi, Keunwoo and Nam, Juhan},
  journal={arXiv preprint arXiv:2510.01698},
  year={2025}
}

@article{tian2025mucpt,
  title={MuCPT: Music-related Natural Language Model Continued Pretraining},
  author={Tian, Kai and Mao, Yirong and Bi, Wendong and Wang, Hanjie and Wenhui, Que},
  journal={arXiv preprint arXiv:2511.14245},
  year={2025}
}

@article{baichuan_finance,
  title={Baichuan4-finance technical report},
  author={Zhang, Hanyu and Qiu, Boyu and Feng, Yuhao and Li, Shuqi and Ma, Qian and Zhang, Xiyuan and Ju, Qiang and Yan, Dong and Xie, Jian},
  journal={arXiv preprint arXiv:2412.15270},
  year={2024}
}

@article{distilqwen,
  title={DistilQwen2. 5: Industrial Practices of Training Distilled Open Lightweight Language Models},
  author={Wang, Chengyu and Yan, Junbing and Yue, Yuanhao and Huang, Jun},
  journal={arXiv preprint arXiv:2504.15027},
  year={2025}
}

@article{li2017learning,
  title={Learning without forgetting},
  author={Li, Zhizhong and Hoiem, Derek},
  journal={IEEE transactions on pattern analysis and machine intelligence},
  volume={40},
  number={12},
  pages={2935--2947},
  year={2017},
  publisher={IEEE}
}

@article{zhang2024map,
  title={Map-neo: Highly capable and transparent bilingual large language model series},
  author={Zhang, Ge and Qu, Scott and Liu, Jiaheng and Zhang, Chenchen and Lin, Chenghua and Yu, Chou Leuang and Pan, Danny and Cheng, Esther and Liu, Jie and Lin, Qunshu and others},
  journal={arXiv preprint arXiv:2405.19327},
  year={2024}
}

@article{bai2024mt,
  title={Mt-bench-101: A fine-grained benchmark for evaluating large language models in multi-turn dialogues},
  author={Bai, Ge and Liu, Jie and Bu, Xingyuan and He, Yancheng and Liu, Jiaheng and Zhou, Zhanhui and Lin, Zhuoran and Su, Wenbo and Ge, Tiezheng and Zheng, Bo and others},
  journal={arXiv preprint arXiv:2402.14762},
  year={2024}
}

@article{prox_example,
  title={Programming every example: Lifting pre-training data quality like experts at scale},
  author={Zhou, Fan and Wang, Zengzhi and Liu, Qian and Li, Junlong and Liu, Pengfei},
  journal={arXiv preprint arXiv:2409.17115},
  year={2024}
}

@article{reversal_curse,
  title={The Reversal Curse: LLMs trained on" A is B" fail to learn" B is A"},
  author={Berglund, Lukas and Tong, Meg and Kaufmann, Max and Balesni, Mikita and Stickland, Asa Cooper and Korbak, Tomasz and Evans, Owain},
  journal={arXiv preprint arXiv:2309.12288},
  year={2023}
}

@article{qurating,
  title={Qurating: Selecting high-quality data for training language models},
  author={Wettig, Alexander and Gupta, Aatmik and Malik, Saumya and Chen, Danqi},
  journal={arXiv preprint arXiv:2402.09739},
  year={2024}
}

@article{fineweb,
  title={The fineweb datasets: Decanting the web for the finest text data at scale},
  author={Penedo, Guilherme and Kydl{\'\i}{\v{c}}ek, Hynek and Lozhkov, Anton and Mitchell, Margaret and Raffel, Colin A and Von Werra, Leandro and Wolf, Thomas and others},
  journal={Advances in Neural Information Processing Systems},
  volume={37},
  pages={30811--30849},
  year={2024}
}

@article{demify_cpt,
  title={Demystifying domain-adaptive post-training for financial llms},
  author={Ke, Zixuan and Ming, Yifei and Nguyen, Xuan-Phi and Xiong, Caiming and Joty, Shafiq},
  journal={arXiv preprint arXiv:2501.04961},
  year={2025}
}

@article{lin2024rho,
  title={Rho-1: Not all tokens are what you need},
  author={Lin, Zhenghao and Gou, Zhibin and Gong, Yeyun and Liu, Xiao and Shen, Yelong and Xu, Ruochen and Lin, Chen and Yang, Yujiu and Jiao, Jian and Duan, Nan and others},
  journal={arXiv preprint arXiv:2404.07965},
  year={2024}
}

@article{wei2024measuring,
  title={Measuring short-form factuality in large language models},
  author={Wei, Jason and Karina, Nguyen and Chung, Hyung Won and Jiao, Yunxin Joy and Papay, Spencer and Glaese, Amelia and Schulman, John and Fedus, William},
  journal={arXiv preprint arXiv:2411.04368},
  year={2024}
}

@article{nguyen2024ceval,
  title={CEval: A benchmark for evaluating counterfactual text generation},
  author={Nguyen, Van Bach and Schl{\"o}tterer, J{\"o}rg and Seifert, Christin},
  journal={arXiv preprint arXiv:2404.17475},
  year={2024}
}

@inproceedings{li2024cmmlu,
  title={Cmmlu: Measuring massive multitask language understanding in chinese},
  author={Li, Haonan and Zhang, Yixuan and Koto, Fajri and Yang, Yifei and Zhao, Hai and Gong, Yeyun and Duan, Nan and Baldwin, Timothy},
  booktitle={Findings of the Association for Computational Linguistics: ACL 2024},
  pages={11260--11285},
  year={2024}
}

@article{Yang2024Qwen25TR,
  title={Qwen2.5 Technical Report},
  author={Qwen An Yang and Baosong Yang and Beichen Zhang and Binyuan Hui and Bo Zheng and Bowen Yu and Chengyuan Li and Dayiheng Liu and Fei Huang and Guanting Dong and Haoran Wei and Huan Lin and Jian Yang and Jianhong Tu and Jianwei Zhang and Jianxin Yang and Jiaxin Yang and Jingren Zhou and Junyang Lin and Kai Dang and Keming Lu and Keqin Bao and Kexin Yang and Le Yu and Mei Li and Mingfeng Xue and Pei Zhang and Qin Zhu and Rui Men and Runji Lin and Tianhao Li and Tingyu Xia and Xingzhang Ren and Xuancheng Ren and Yang Fan and Yang Su and Yi-Chao Zhang and Yunyang Wan and Yuqi Liu and Zeyu Cui and Zhenru Zhang and Zihan Qiu and Shanghaoran Quan and Zekun Wang},
  journal={arXiv preprint arXiv:2412.15115},
  year={2024},
}

@article{TALKPLAY,
  publtype={informal},
  author={Seungheon Doh and Keunwoo Choi and Juhan Nam},
  title={TALKPLAY: Multimodal Music Recommendation with Large Language Models},
  year={2025},
  month={February},
  cdate={1738368000000},
  journal={CoRR},
  volume={abs/2502.13713},
  url={https://doi.org/10.48550/arXiv.2502.13713}
}

@inproceedings{melchiorre2025just,
  title={Just ask for music (jam): Multimodal and personalized natural language music recommendation},
  author={Melchiorre, Alessandro B and Epure, Elena V and Masoudian, Shahed and Escobedo, Gustavo and Hausberger, Anna and Moussallam, Manuel and Schedl, Markus},
  booktitle={Proceedings of the Nineteenth ACM Conference on Recommender Systems},
  pages={615--620},
  year={2025}
}

@article{palumbo2025text2tracks,
  title={Text2Tracks: Prompt-based Music Recommendation via Generative Retrieval},
  author={Palumbo, Enrico and Penha, Gustavo and Damianou, Andreas and Garc{\'\i}a, Jos{\'e} Luis Redondo and Heath, Timothy Christopher and Wang, Alice and Bouchard, Hugues and Lalmas, Mounia},
  journal={arXiv preprint arXiv:2503.24193},
  year={2025}
}

@INPROCEEDINGS{11199092,
  author={Sharanarthi, Tanush},
  booktitle={2025 World Skills Conference on Universal Data Analytics and Sciences (WorldSUAS)}, 
  title={AI Agent-Based Framework for Personalized Music Recommendation}, 
  year={2025},
  volume={},
  number={},
  pages={1-5},
  doi={10.1109/WorldSUAS66815.2025.11199092}}

@ARTICLE{11174114,
  author={Wang, Jiarong and Wu, Jiaji and Tan, Mingzhou and Zhu, Lingxuan},
  journal={IEEE Transactions on Computational Social Systems}, 
  title={Emotion-Aware Conversational Music Recommendation With Multiagent System}, 
  year={2025},
  volume={},
  number={},
  pages={1-14},
  doi={10.1109/TCSS.2025.3599008}}

@inproceedings{AgentCF,
author = {Zhang, Junjie and Hou, Yupeng and Xie, Ruobing and Sun, Wenqi and McAuley, Julian and Zhao, Wayne Xin and Lin, Leyu and Wen, Ji-Rong},
title = {AgentCF: Collaborative Learning with Autonomous Language Agents for Recommender Systems},
year = {2024},
isbn = {9798400701719},
publisher = {Association for Computing Machinery},
address = {New York, NY, USA},
url = {https://doi.org/10.1145/3589334.3645537},
doi = {10.1145/3589334.3645537},
booktitle = {Proceedings of the ACM Web Conference 2024},
pages = {3679–3689},
numpages = {11},
location = {Singapore, Singapore},
series = {WWW '24}
}

\appendix
\quad\\
{\LARGE \textbf{Appendix}}
\hypertarget{sec:appendix}{}
\label{sec:appendix}
\section{Data Construction of MusicCPT for WeMusic-Base}
\hypertarget{sec:cpt_data_appendix}{}
\label{sec:cpt_data_appendix}
We details the primary data sources employed in the continual pretraining process of CPT for WeMusic-Base.

\textbf{Music-related data from open source generic corpora.}
A lightweight music-domain classifier is trained to extract music-related documents from the publicly available pretraining Matrix corpus \cite{zhang2024map}. This corpus integrates a broad spectrum of open-source Chinese and English datasets from diverse sources, such as web crawls, books, question-answer pairs, and Wikipedia. Through this process, we obtain approximately 20B tokens of music-related text (corresponding to roughly 40M samples), which constitutes less than 1\% of the original Matrix corpus. These data serve as foundational cross-domain and cross-lingual background knowledge for music-related concepts in subsequent modeling stages.

\textbf{WeChat Articles.}
To improve the model's comprehension of individual songs and its capability for text-based song retrieval, we anchor our approach on a collection of high-play-frequency songs from the WeChat Listen platform. For each selected song, we retrieve relevant articles, which are subsequently filtered by the aforementioned music-domain classifier to eliminate irrelevant or too low-quality content. 

\textbf{WeChat Encyclopedia.}
We collect encyclopedia pages categorized under music, entertainment, and related fields, with special emphasis on artist entries. These entries integrate biographical profiles, representative works, stylistic labels, award records, and other structured and unstructured information, collectively forming a foundation for building an internal ``artist–works–style'' knowledge graph within the model.

\textbf{Music Books.} We further collect thousands of music-related books (such as music theory introductions, music history, reviews, and interviews), and convert their PDF files into plain text. These data are more oriented towards long form narrative and systematic knowledge, helping the model acquire deeper understanding of music theory, historical context, and professional critique.

\textbf{Collaborative Data.}
To address the limited personalization capacity of general-purpose large language models, which typically lack collaborative recommendation data grounded in real-world scenarios, we sample user listening behavior sequences from the WeChat Listen platform. In addition, we treat songs that users liked or listened multiple times as anchors, and simulate users asking natural language questions about these anchor songs. This yields training instances conditioned on both user listening preferences and search queries, which further strengthen the model’s ability to model user preference in personalized retrieval and conversational recommendation scenarios.
The training corpus is further augmented with public playlists, each comprising a title, a textual description, and an ordered list of songs. 
The song assemblages within playlists inherently encapsulate rich collaborative-filtering signals and thematic structures, such as those organized by mood, scenario, or genre, which substantially enhance the model's capacity for list‑wise song recommendation. 

\textbf{Music Metadata.}
To enhance the model's factual memory capacity, we curate metadata for widely popular songs, encompassing titles, artists, lyrics, genres, languages, albums and their descriptions, creation backgrounds, BPM (beats per minute), and related attributes. Additionally, we integrate weakly labeled multimodal tags, such as emotion, style, instrument, and rhythm categories. They are predicted by content understanding models. The combination of such structured and semi-structured metadata is crucial for establishing robust mappings among musical entities, attributes, and semantic tags during CPT.

\textbf{Song Description Snippets.}
Building upon the WeChat articles and user comment data described earlier, we first employ a powerful large language model (LLM) to filter article paragraphs and extract segments closely associated with specific songs. These segments are then condensed into high‑quality, song‑level descriptive snippets, as illustrated in Fig.~\ref{fig:bi_aug}. In parallel, we gather billions of user comments on songs and similarly apply a strong LLM to filter out irrelevant or information‑sparse remarks, as well as to generate aggregated summaries encapsulating emotions, scenarios, narrative contexts, and listening impressions.

\section{Evaluation of Conversational Capability}
\label{sec:appendix_eval_conversation}
We design a comprehensive set of evaluation metrics tailored for Listen, covering aspects such as intent understanding, response quality, and task-oriented effectiveness. Listen consists of multi-turn dialogues. Each Listen response comprises three key components: an intention (wrapped within <intention>tags), the response text (within <text> tags), and optionally a specific music recommendation (within <music> tags). The evaluation framework assesses all three components. We designed a set of 9 metrics to evaluate the Listen system. The first three metrics are adapted from MT-Bench-101 \cite{bai2024mt}.

\textbf{Context Memory, CM}. 
This metric quantifies the system's proficiency in maintaining long-range conversational coherence by accurately recalling and leveraging information from previous dialogue turns. This capability is crucial for natural, engaging multi-turn interactions. 

\textbf{Anaphora Resolution, AR}.
This metric assesses the dialogue system's capability to resolve anaphora and maintain referential coherence, which is critical for seamless multi-turn conversations. Anaphora here refers to linguistic expressions (e.g., pronouns, demonstratives) whose interpretation depends on antecedent context. 

\textbf{Topic Shift, TS}.
The user frequently switches between discussing different singers or songs without any clear transition. This metric quantifies the dialogue system's agility in detecting user-initiated topic shifts and adapting its responses accordingly, which is crucial for maintaining natural and engaging multi-turn conversations. 

\textbf{Intention Understanding, IU}.
The accuracy of intent understanding for the "like/collect song" domain is quantified using a discrete scoring system (scores: 1, 2, 10), based on the alignment between the user's query and the system's response. The evaluation logic is formally defined as a two-step decision process:
First, the user's query is analyzed to determine the presence or absence of an intent to "like" or "collect" the current song, as well as other intents. Second, the score is then assigned based on the criteria.

\textbf{Contextual Recommendation, CR}.
This metric assesses whether a song recommendation is contextually appropriate to the dialogue flow, with a specific focus on penalizing intrusive recommendations that break conversational coherence.

\textbf{Response Non-Repetitiveness, RN}.
This metric evaluates the conciseness and information efficiency of the system's responses by penalizing unnecessary repetition of ideas or phrasing across dialogue turns, especially when the same song is recommended repeatedly in near-consecutive turns. The evaluation is specifically triggered for responses tagged with the <intention> \(SONG\_SEARCH\) </intention>intent.

\textbf{Functional Boundary, FB}.
This metric evaluates the system's capability to recognize and honestly communicate its functional limitations, a critical factor for user trust and interaction safety. Performance is scored on a scale of 1 to 10, with the core principle being the minimization of "hallucination" (i.e., generating factually incorrect or ungrounded information).

\textbf{Feedback Handling, FH}.
This metric measures the system's proficiency in handling user feedback that critiques the accuracy or appropriateness of a previous system utterance. The evaluation is triggered specifically by user turns containing corrective intent. 

\textbf{Probing Questions, PQ}.
This metric measures the system's proficiency in employing clarifying and guiding questions to manage conversational ambiguity and enhance interaction depth. It is assessed along three primary dimensions: 1) Sensitivity in recognizing the need for clarification, 2) Proactive guidance in maintaining conversational momentum, and 3) Quality of question design. Note: This metric is relaxed for straightforward command-execution turns (e.g., "play this song").

\begin{figure*}[h]
\centering
\includegraphics[width=0.5\textwidth]{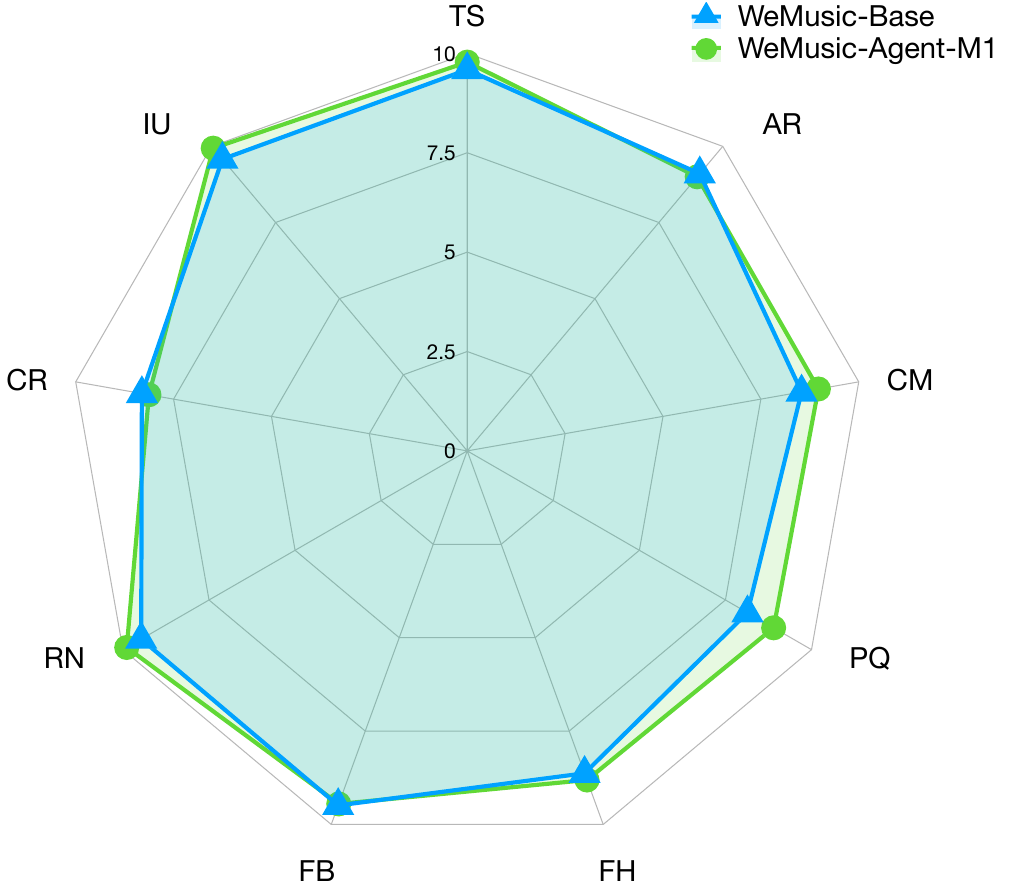}
\caption{Radar chart of conversational capability evaluation.}
\label{fig:score_radar_chart}
\end{figure*}

Figure~\ref{fig:score_radar_chart} presents a comparative analysis of conversational capabilities between the two models using radar charts.

\section{Experiments on different data mixture ratios in CPT}
\label{mixture_section}
\begin{figure}[htbp]
    \centering
    \begin{subfigure}[b]{0.48\textwidth}
        \includegraphics[width=\linewidth]{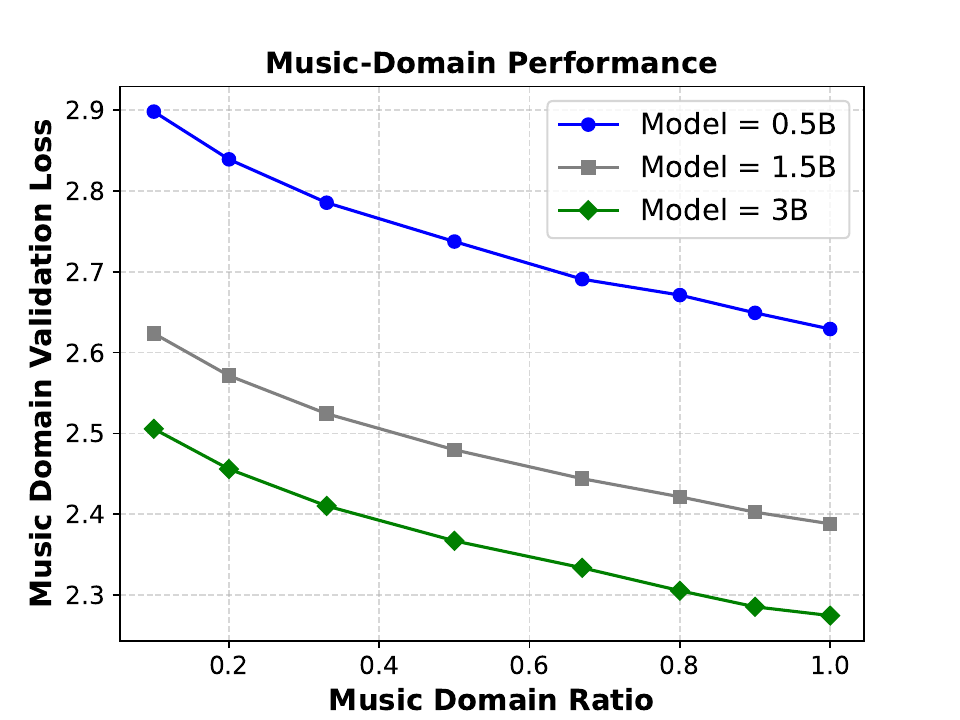}
        \caption{The music domain performances with different music domain ratio.}
        \label{fig:sub1}
    \end{subfigure}
    \hfill 
    \begin{subfigure}[b]{0.48\textwidth}
        \includegraphics[width=\linewidth]{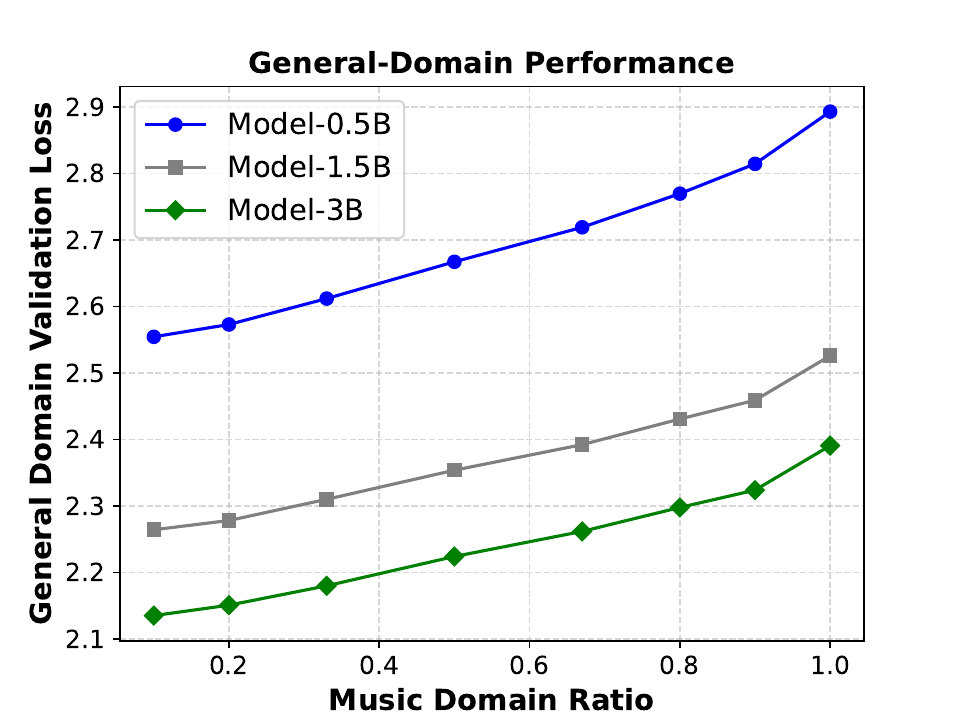} 
        \caption{The general domain performances with different music domain ratio.}
        \label{fig:sub2}
    \end{subfigure}
    \caption{Experiments to tune mixture ratio between music and general domain.}
    \label{fig:mixture_curve}
\end{figure}

Figure~\ref{fig:mixture_curve} plots the validation loss on the music domain dev set and a held out general domain dev set for different mixture ratios.  As the proportion of music data increases from $0.1$ to $1.0$, the music domain validation loss monotonically decreases, whereas the general domain validation loss gradually increases, forming a clear trade off: more music data helps the model better capture music knowledge and stylistic nuances, but slightly harms its general capability. 


\subsection{Details on the Metric Scores} 
Performance is scored on a scale of 1 to 10 based on the following criteria for each metric.

\textbf{Context Memory, CM}. 
\begin{itemize}
\item Default Case: If the current user utterance requires no reference to dialogue history, the system is not penalized for not recalling it.
\item Score 10 (Excellent): Recall and integration of historical details are flawless. The system seamlessly weaves prior information into the current response, significantly enhancing clarity and continuity.
\item Scores 7-9 (Good): The system stably recalls and integrates historical information to maintain dialogue coherence. Responses are largely consistent and contextually appropriate, with only minor, non-disruptive lapses.
\item Scores 4-6 (Fair): The system shows partial memory, occasionally referencing prior context, but with inconsistencies or omissions that lead to minor incoherencies.
\item Scores 1-3 (Poor): Responses demonstrate a severe lack of contextual memory, leading to contradictions, non-sequiturs, or a complete disregard for established information, which disrupts the dialogue flow.
\end{itemize}

\textbf{Anaphora Resolution, AR}.
\begin{itemize}
\item Default Case:  If the user's utterance contains no anaphoric expressions, a score of 10 is assigned by default.
\item Score 10 (Excellent):  Awarded for flawless anaphora resolution, where the response demonstrates a perfect understanding and integration of the referential information, resulting in a highly coherent and contextually precise answer.
\item Scores 7-9 (Good):  Awarded for a strong anaphora resolution capability. Responses are largely accurate and coherent, with only minor imperfections in contextual linkage. Crucially, this score range also applies when the system correctly identifies a user request as unfulfillable (e.g., a hardware control command like "turn up the volume") and responds with a clear statement of its inability (e.g., "I don't know how to do that"). This is considered a valid resolution of the pragmatic intent behind the anaphor.
\item Scores 4-6 (Fair):  Awarded when the system demonstrates a partial understanding of the anaphora but the response contains noticeable inaccuracies or fails to fully leverage the contextual information.
\item Scores 1-3 (Poor):  Awarded when the system fails to identify or correctly interpret the anaphoric reference, leading to a response that is inaccurate or completely disconnected from the prior dialogue context.
\end{itemize}

\textbf{Topic Shift, TS}.
\begin{itemize}
\item Score 10 (Excellent):  The system handles the topic shift flawlessly. The transition is smooth, and the response is fully committed to the new topic, providing a contextually perfect and engaging reply without any irrelevant carry-over.
\item Scores 7-9 (Good):  The system reliably detects the topic shift and engages with the new topic appropriately. Responses are largely relevant and coherent, with minimal to no interference from the prior conversation.
\item Scores 4-6 (Fair):  The system demonstrates a basic ability to follow the new topic but its responses may occasionally contain residual elements from the previous discussion, indicating incomplete adaptation and leading to partially relevant replies.
\item Scores 1-3 (Poor):  The system fails to detect the topic shift, persistently reverting to the previous topic or producing responses that are heavily influenced by prior context, resulting in irrelevance or confusion.
\end{itemize}

\textbf{Intention Understanding, IU}.
\begin{itemize}
\item Score 10 (Correct): Awarded in two scenarios: a) the user's intent is absent and the system's response does not contain the tag like <intention> \(LIKE\) </intention>; or b) the user's intent is present and the system's response correctly contains the tag.
\item Score 2 (Incorrect): Awarded when the user's intent is absent, but the system's response incorrectly contains the tag like <intention> \(LIKE\) </intention> (a False Positive).
\item Score 1 (Incorrect): Awarded when the user's intent is present, but the system's response either contains an incorrect intention tag (e.g., <intention> \(PLAYLIST\_LIKE\) </intention>) or lacks the correct tag entirely (a False Negative/Misclassification).
\end{itemize}

\textbf{Contextual Recommendation, CR}.
\begin{itemize}
\item Score 10 (Excellent): Awarded in three scenarios: a) The dialogue contains no song recommendation. b) A recommendation is highly coherent with the conversation (e.g. following a user's statement of preference). c) A recommendation serves as a graceful fallback when the user's primary request (e.g., checking the weather) exceeds the system's functional boundaries.
\item Score 7-9 (Good): Awarded when a recommendation is provided in direct response to a clear user request (e.g., "Recommend a song") and is correctly tagged with a search intent (<intention> \(SONG\_SEARCH\) </intention>).
\item Score 4-6 (Fair): Awarded when a specific song is recommended, but the system explicitly seeks user consent before playing (e.g., using phrases like "Play this song?"), thus mitigating intrusiveness.
\item Score 1-3 (Poor): Awarded when the system proactively recommends a specific song while the user is discussing the content or emotion of the currenttrack without any explicit request, and the recommendation is tagged with an intent (e.g., <intention> \(MUSIC\_CHAT\) </intention>) misaligned with a direct search action.
\end{itemize}

\textbf{Response Non-Repetitiveness, RN}.
\begin{itemize}
\item Score 10 (Default): Awarded automatically for any response whose intention is not "\(SONG\_SEARCH\)", as the metric is designed specifically for this action-oriented intent.
\item Score 6-10 (Concise and Novel): Awarded for a "\(SONG\_SEARCH\)" response where the textual content (<text>) introduces new information or descriptions compared to the previous turn, demonstrating efficient communication.
\item Score 1-5 (Redundant and Repetitive): Awarded for a "\(SONG\_SEARCH\)" response where the textual content exhibits substantial lexical or semantic overlap with the content of the previous system turn. This indicates a failure to provide novel information, resulting in verbosity.
\end{itemize}

\textbf{Functional Boundary, FB}.
\begin{itemize}
\item Scores 10 (Excellent): Awarded for optimal responses where the system not only correctly declines the request but also proactively and smoothly redirects the conversation to a music-related topic within its capabilities, thereby maintaining user engagement while respecting its functional boundaries. For example, 
Q: "Remind me to take out the trash in 30 minutes, will you?"
A: "Oh, I can't actually set alarms myself—you'll have to take care of that! But once you're done, I can hang out and listen to some music with you~"
\item Scores 7-9 (Good): The baseline score for correctly identifying an unsupported request and explicitly declining to fulfill it (e.g., "I cannot do that").
\item Scores 4-6 (Fair): Assigned to responses that partially address the out-of-scope request but still contain ambiguities or minor inaccuracies that could potentially mislead the user. The score within this range corresponds to the degree of hallucination risk reduction.
\item Scores 1-3 (Poor): Awarded when the system provides a factually incorrect response by falsely claiming to fulfill a request that is beyond its defined capabilities (e.g., agreeing to set an alarm or report the weather), indicating a severe failure in self-awareness.
\end{itemize}

\textbf{Feedback Handling, FH}.
\begin{itemize}
\item Default Case: If the user's utterance does not constitute a critique or point out a potential error, a default score of 8 is assigned, as the metric is not applicable.
\item Score 10 (Excellent): The system executes a perfect response: it politely acknowledges the feedback, clearly states whether the original answer was incorrect (providing a flawless correction) or correct (providing compelling reasoning to uphold it), and maintains a confident yet courteous tone throughout.
\item Scores 7-9 (Good): The system accurately acknowledges the critique and provides a substantive correction if an error is verified. The updated response is largely accurate and addresses the user's concern directly.
\item Scores 4-6 (Fair): The system recognizes the critique and attempts a response. However, the correction may be only partial, the updated information may still be flawed, or the system's stance is non-committal (e.g., using "perhaps" or "maybe"), demonstrating low confidence.
\item Scores 1-3 (Poor):The system fails to recognize the user's critique or completely ignores it, proceeding as if no feedback was given. Alternatively, it may acknowledge the critique but fail to provide a necessary correction or explanation.
\end{itemize}

\textbf{Probing Questions, PQ}.
\begin{itemize}
\item Score 10 (Excellent): The system demonstrates mastery by asking anticipatory questions to prevent ambiguity. Its questioning and answering form a complete "cognitive closed-loop," with multi-layered topic nesting and a personalized style, significantly deepening the engagement.
\item Scores 7-9 (Good): The system quickly identifies ambiguity and uses precise, often sequential questions to efficiently clarify user intent. Responses are systematic, and the dialogue remains natural and coherent, demonstrating good conversational guidance.
\item Scores 4-6 (Fair): The system recognizes the need for clarification but its questions are imprecise (e.g., overly open-ended). Responses partially address the clarified points, but the questioning rhythm may be awkward, showing inconsistent balance between guidance and user lead.
\item Scores 1-3 (Poor): The system fails to ask necessary questions, instead guessing the intent of ambiguous queries. Any questions asked are disruptive, and responses remain incomplete or illogical. Topic shifts are abrupt.
\end{itemize}

\end{document}